\documentclass{article}

\PassOptionsToPackage{numbers, compress}{natbib}
\usepackage[preprint]{neurips_2025}

\usepackage[utf8]{inputenc} % allow utf-8 input
\usepackage[T1]{fontenc}    % use 8-bit T1 fonts
\usepackage{hyperref}       % hyperlinks
\usepackage{url}            % simple URL typesetting
\usepackage{booktabs}       % professional-quality tables
\usepackage{amsfonts}       % blackboard math symbols
\usepackage{nicefrac}       % compact symbols for 1/2, etc.
\usepackage{microtype}      % microtypography
\usepackage{xpatch}
\usepackage{amsmath} % Add this line to include the amsmath package
\usepackage{graphicx}       % for including figures
\usepackage{tabularx}
\usepackage{arydshln}
\usepackage{subcaption}
\usepackage{multirow}
\usepackage{multicol}

\usepackage{xpatch}
\usepackage{makecell}
\usepackage{colortbl}
\usepackage[normalem]{ulem}
\usepackage[table,svgnames]{xcolor}
\usepackage{nicematrix}
\usepackage{tcolorbox}
\usepackage{titlesec}
\usepackage[font=small]{caption}

\definecolor{lightblue}{rgb}{0.88, 0.92, 1} % 定义淡蓝色

\makeatletter
\xapptocmd{\NAT@bibsetnum}{\setlength{\leftmargin}{0pt}\setlength{\itemindent}{\labelwidth}\addtolength{\itemindent}{\labelsep}}{}{}
\makeatother

\titlespacing{\paragraph}{%
  0pt}{%              left margin
  0.01 \baselineskip}{% space before (vertical)
  1em}% 

\newcommand{\wlc}[2]{\bgroup\textcolor{cyan}{\sout{#1} #2}\egroup}

\newcommand{\dss}{\text{DeepSeek-R1-Distill-Qwen-1.5B}}
\newcommand{\dsm}{\text{DeepSeek-R1-Distill-Qwen-7B }}
\newcommand{\dsl}{\text{DeepSeek-R1-Distill-Qwen-32B }}
\newcommand{\mone}{\textsc{Laser}}
\newcommand{\mtwo}{\textsc{Laser-D}}
\newcommand{\mthree}{\textsc{Laser-DE}}

\newcommand{\csymbol}{C(y)}

\newcommand{\jhc}[2]{\bgroup\textcolor{magenta}{\sout{#1} #2}\egroup}

\title{Learn to Reason Efficiently with \\ Adaptive Length-based Reward Shaping}
\author{%
  \textbf{Wei Liu}\textsuperscript{1}\thanks{Correspondence to Wei Liu (wliucn@cse.ust.hk) and Junxian He (junxianh@cse.ust.hk)} \quad
  \textbf{Ruochen Zhou}\textsuperscript{2} \quad
  \textbf{Yiyun Deng}\textsuperscript{1} \quad
  \textbf{Yuzhen Huang}\textsuperscript{1} \quad
  \textbf{Junteng Liu}\textsuperscript{1} \\
  \textbf{Yuntian Deng}\textsuperscript{3} \quad
  \textbf{Yizhe Zhang}\textsuperscript{4} \quad
  \textbf{Junxian He}\textsuperscript{1}\footnotemark[1] \\
  \textsuperscript{1}The Hong Kong University of Science and Technology \quad
  \textsuperscript{2}City University of Hong Kong \\
  \textsuperscript{3}University of Waterloo \quad
  \textsuperscript{4}Apple \\
}

\begin{document}

\maketitle

% \begin{figure}[htbp]
%     \centering
%     \begin{subfigure}[t]{0.48\textwidth}
%         \centering
%         \includegraphics[width=\textwidth]{figures/average_performance.pdf}
%         % \caption{Accuracy and Response Length on All Benchmarks (MATH500, AIME2024, AMC2023, Olympiad Bench)}
%         \label{fig:pareto-avg}
%     \end{subfigure}
%     \hfill
%     \begin{subfigure}[t]{0.48\textwidth}
%         \centering
%         \includegraphics[width=\textwidth]{figures/average_aime.pdf}
%         % \caption{Accuracy and Response Length on AIME2024}
%         \label{fig:pareto-aime}
%     \end{subfigure}
%     \caption{\jh{on the legend, our methods should be put at bottom, and more vallina baselines are on the top}Accuracy and response length across various methods. (a) accuracy and response Length on all benchmarks (MATH500, AIME2024, AMC2023, Olympiad Bench). (b) accuracy and response length on AIME2024. Each point represents a single training run with different hyper-parameters. Given the high computational cost of obtaining this figure, the base model used is \dss. Our methods, DARL-E, DARL, and RTL, achieve a Pareto-optimal trade-off compared to all other methods.\jh{these two figures are kinda repetitive, i think it is better to replace one subfigure with a case study example}}
%     \label{fig:pareto}
% \end{figure}

\begin{abstract}
% \jh{changed the title, ``difficulty-aware RL'' sounds like a boring term, it reflects neither length reward nor dynamic nature}
  Large Reasoning Models (LRMs) have shown remarkable capabilities in solving complex problems through reinforcement learning (RL), particularly by generating long reasoning traces. However, these extended outputs often exhibit substantial redundancy, which limits the efficiency of LRMs. In this paper, we investigate RL-based approaches to promote reasoning efficiency.
  Specifically, we first present a unified framework that formulates various efficient reasoning methods through the lens of length-based reward shaping. Building on this perspective, we propose a novel \textbf{L}ength-b\textbf{A}sed \textbf{S}t\textbf{E}p \textbf{R}eward shaping method (\mone), which employs a step function as the reward based on target length. \mone{} surpasses previous methods, achieving a superior Pareto-optimal balance between performance and efficiency.
  Next, we further extend \mone{} based on two key intuitions: (1) The reasoning behavior of the model evolves dynamically during training, necessitating reward specifications that are also adaptive and dynamic; (2) Rather than uniformly encouraging shorter or longer chains of thought (CoT), we posit that length-based reward shaping should be difficulty-aware i.e., it should penalize lengthy CoTs more for easy queries. This approach is expected to facilitate a combination of fast and slow thinking, leading to a better overall tradeoff. The resulting method is termed \mtwo{} (Dynamic and Difficulty-aware).
  % Then we formulate this baseline as a length-based reward function, and propose a unified view to formulate and compare various length-based reward shaping approaches. 
  % Based on this unified view, we first propose a new length-based reward shaping approach, propose three new length-based reward shaping techniques that all outperform previous approaches, achieving superior Pareto-optimal between performance and efficiency. Notably,  between 
  % we explore RL-based chain-of-thought (CoT) compression. We propose a unified view for existing length reward-shaping methods and connect the truncation method, a simple yet effective baseline, to this view. Building on this view, we propose a novel reward-shaping method, \textbf{R}eward with \textbf{T}arget \textbf{L}ength (RTL), that employs a step funlction guided by a target length. 
  % To further optimize the effectivess-efficiency trade-off, we further propose a \textbf{D}ynamic and \textbf{D}ifficulty-\textbf{A}ware \textbf{L}ength-based \textbf{S}tep \textbf{R}eward shaping (\mtwo) that tailors the target length based on the question's difficulty. 
  Experiments on \dss, \dsm, and \dsl demonstrate that our approach significantly enhances both reasoning performance and response length efficiency. For instance, \mtwo{} and its variant achieve a $+\bf{6.1}$ improvement on AIME2024 while reducing token usage by $\bf{63}$\%. Further analysis reveals that our RL-based compression produces more concise reasoning patterns with less redundant ``self-reflections''. All resources (Models, Code, Data) are available at \url{https://github.com/hkust-nlp/Laser}.
\end{abstract}

\section{Introduction}
\label{sec:intro}

% \begin{figure}
%     \centering
%     \includegraphics{}
%     \caption{Caption}
%     \label{fig:enter-label}
% \end{figure}

\begin{figure}[htbp]
    \centering
    % \begin{subfigure}[t]{0.48\textwidth}
    %     \centering
    %     \includegraphics[width=\textwidth]{figures/average_performance.pdf}
    %     % \caption{Accuracy and Response Length on All Benchmarks (MATH500, AIME2024, AMC2023, Olympiad Bench)}
    %     \label{fig:pareto-avg}
    % \end{subfigure}
    % \hfill
    \begin{subfigure}[c]{0.43\textwidth}
        \centering
        \includegraphics[width=\textwidth]{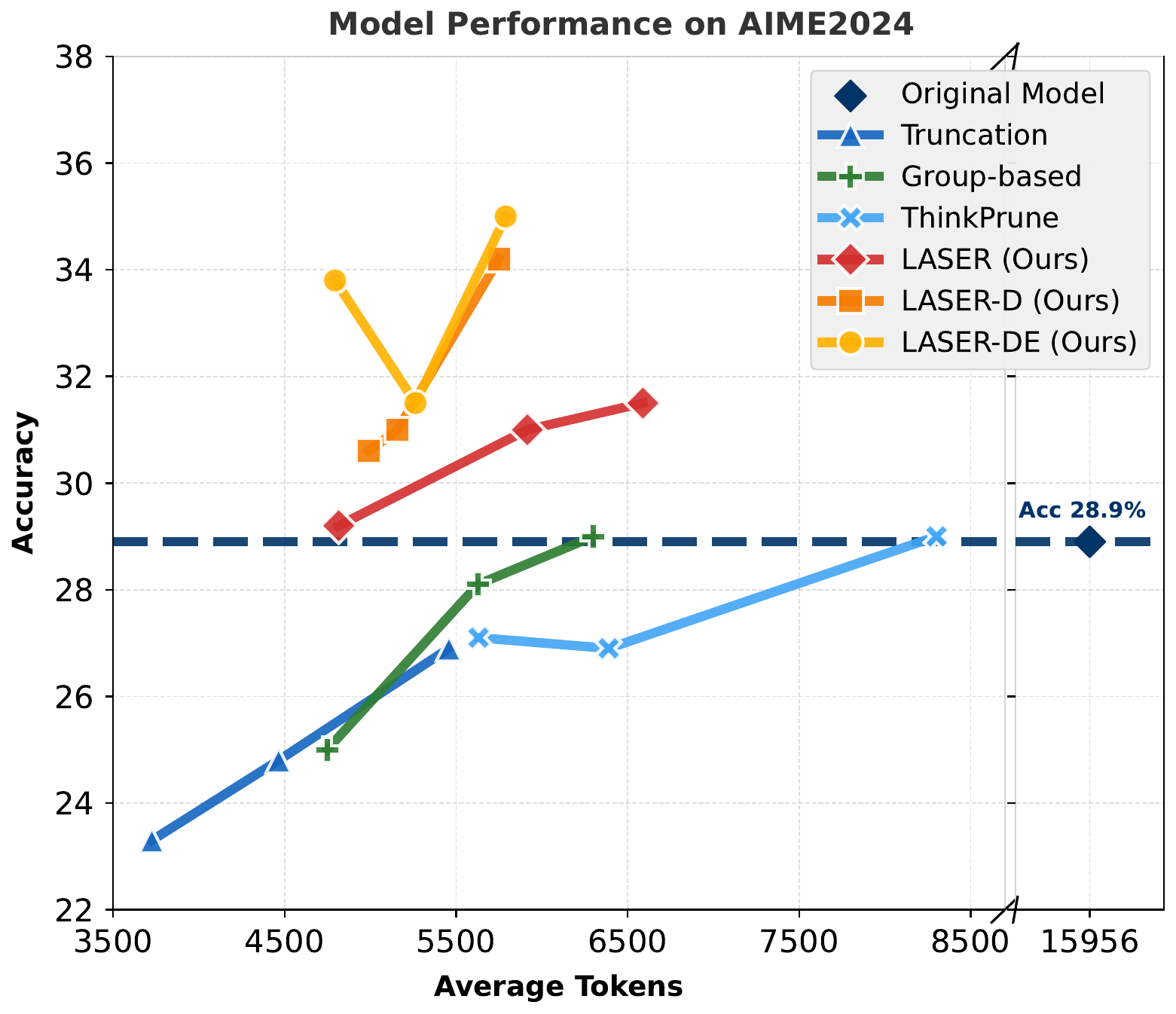}
        % \caption{Accuracy and Response Length on AIME2024}
        \label{fig:pareto-aime}
    \end{subfigure}
    \hfill
    \begin{subfigure}[c]{0.56\textwidth}
      \centering
      \includegraphics[width=\textwidth]{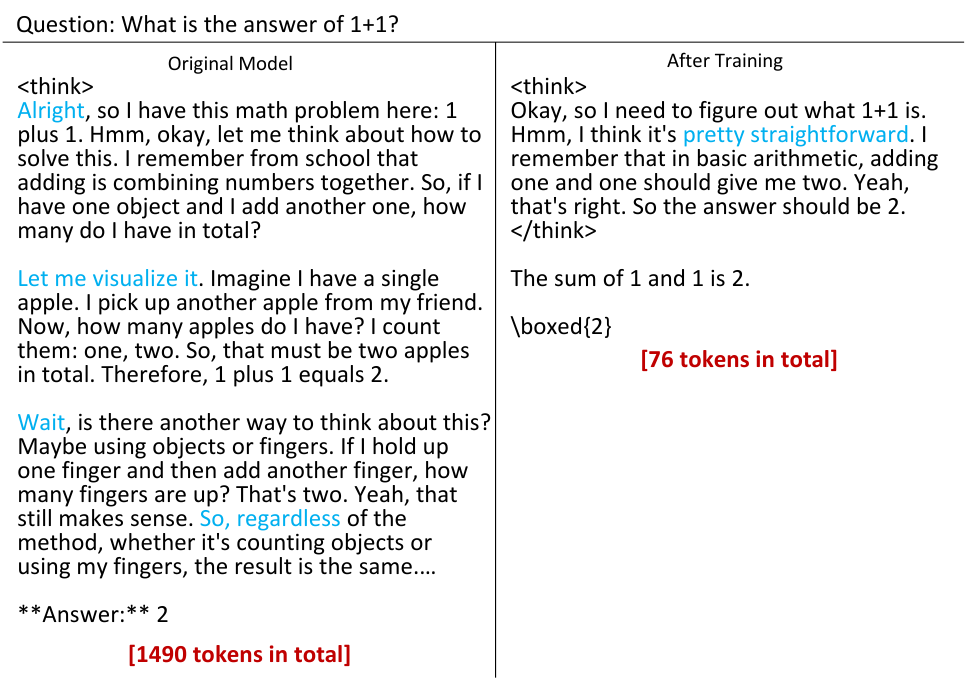}
      % \caption{Accuracy and Response Length on AIME2024}
      \label{fig:case_study}
  \end{subfigure}
            \vspace{-0.65cm}
            \caption{\textbf{Left}: Accuracy and response length on AIME2024. For the figure of more benchmarks, please refer to Appendix~\ref{ap:full_pareto_figures}. Each point represents a single training run with different hyper-parameters. Given the high computational cost of obtaining this figure, the base model used is \dss. Results on 7B and 32B models are in \S\ref{sec:larger_sizes}. Our methods, \mone, \mtwo, and \mthree{} achieve a Pareto-optimal trade-off compared to all other methods. Notably, they yield a $+\bf{6.1}$ improvement in accuracy while reducing tokens usage by $\bf{63}\%$ compared to the original model. \textbf{Right}: Example of a reasoning process after \mtwo{} training. While the original model produces meaningless ``self-reflection'' repeatedly for trivial questions like ``1+1=?'', LRMs after \mtwo{} training efficiently recognize such questions during thinking and provide the answer directly.}
      \label{fig:pareto}
      \vspace{-15pt}
  \end{figure}

  % Large Language Models (LLMs)~\citep{deepseekai2025deepseekv3technicalreport,qwen2025qwen25technicalreport,grattafiori2024llama3herdmodels} have exhibited remarkable capabilities across a broad range of tasks.
%   \jh{citation style, better order the references in terms of last names of the first author (which is the case for most papers), rather than their cited order }
  Recent advancements leveraging reinforcement learning (RL)~\citep{deepseekai2025deepseekr1incentivizingreasoningcapability,kimiteam2025kimik15scalingreinforcement,openai2024openaio1card,zeng2025simplerlzooinvestigatingtamingzero}  demonstrate that LLMs can evolve into powerful Large Reasoning Models (LRMs), capable of producing extended chains of thought (CoT) to enhance their reasoning abilities. However, these longer reasoning trajectories come at the cost of increased token usage and potentially incorporate more compounding errors. Many of the generated tokens tend to be unnecessarily verbose. For example, LRMs may output thousands of tokens to solve elementary math problems that could otherwise be addressed within just a few hundred tokens, as shown in Figure~\ref{fig:pareto} (right). This phenomenon is commonly referred to as the \emph{over-thinking} issue~\citep{chen2025think23overthinkingo1like}.

Previous work~\citep{qu2025surveyefficientreasoninglarge,wang2025harnessingreasoningeconomysurvey} has explored various approaches to improving reasoning efficiency in LRMs, including continuous chain-of-thought (CoT) reasoning~\citep{hao2024traininglargelanguagemodels,ruan2025reasoninglearnlatentthoughts}, supervised fine-tuning (SFT)~\citep{xia2025tokenskipcontrollablechainofthoughtcompression,munkhbat2025selftrainingelicitsconcisereasoning}, and reinforcement learning~\citep{kimiteam2025kimik15scalingreinforcement,arora2025traininglanguagemodelsreason,aggarwal2025l1controllinglongreasoning,hou2025thinkprunepruninglongchainofthought}. Typically, substantial reductions in token usage are accompanied by significant decreases in reasoning accuracy, suggesting a trade-off between efficiency and reasoning performance. Recently, RL-based approaches have demonstrated the most favorable balance between token efficiency and accuracy~\citep{kimiteam2025kimik15scalingreinforcement,arora2025traininglanguagemodelsreason,aggarwal2025l1controllinglongreasoning,hou2025thinkprunepruninglongchainofthought}. 
In this paper, we study RL-based CoT compression, beginning with a simple baseline that yields surprisingly effective results (\S\ref{sec:truncation}). Specifically, we further train long CoT reasoning models using RL with a rule-based correctness reward, while restricting the context window size to a smaller value than the model's typical generated length, so that long responses will be truncated. This approach can substantially improve token efficiency with only a modest reduction in accuracy. 
% Intuitively, this baseline penalizes responses that exceed the context window by truncating them and treating the truncated outputs as incorrect.
% While this simple truncation baseline does not involve the careful design of new length-dependent reward functions as in prior work \jh{cite}, we find that its performance is surprisingly superior to some specially curated reward shaping approaches. 
To better understand the effectiveness of this truncation method and to connect it with other RL approaches that incorporate length-based rewards~\citep{kimiteam2025kimik15scalingreinforcement,arora2025traininglanguagemodelsreason,aggarwal2025l1controllinglongreasoning}, we introduce a unified length-based reward shaping perspective that encompasses various RL strategies for mitigating overthinking (\S\ref{sec:unified}). Building on this reward shaping formulation, we extend the truncation approach as a novel reward shaping method (\S\ref{sec:lsr}) that employs a step function as the reward guided by a desired target length. We refer to this approach as \emph{\mone} (\textbf{L}ength-b\textbf{A}sed \textbf{S}t\textbf{E}p \textbf{R}eward). \mone{} achieves the best trade-off between reasoning performance and efficiency among all evaluated baselines.

Next, we identify two key points that are lacking in \mone{}: (1) the desired reasoning length should evolve during training as the model's reasoning behaviors dynamically change, and (2) rather than uniformly encouraging short or long CoT, length-based reward shaping should be \emph{difficulty-aware} -- allowing harder questions a higher token limit while constraining easier questions to fewer tokens. To this end, we propose a \textbf{D}ynamic and \textbf{D}ifficulty-aware \textbf{L}ength-b\textbf{A}sed \textbf{S}t\textbf{E}p \textbf{R}eward for RL (\mtwo{}), which adaptively applies different length reward shaping based on problem difficulty. Notably, our algorithm is fully automated with an integrated automatic adapting module, eliminating the need for manual procedural tuning. We also introduce a variant of \mtwo{}, called \mthree{}, which explicitly encourages additional exploration on incorrect responses, enabling models to discover potentially correct reasoning patterns through extended deliberation.
% y measuring the difficulty of each question and automatically assigning the optimal target length for each difficulty level. 
% Notably, our algorithm is entirely automated with a monitoring module, without manual tuning of the specific procedures. 
% In addition, to further enhance the exploration capability of policy models and improve reasoning for initially incorrect responses, we introduce a variant called \emph{DARL-E}. Inspired by findings that longer reasoning can benefit model performance~\citep{muennighoff2025s1simpletesttimescaling,hou2025thinkprunepruninglongchainofthought}, and that incorrect outputs tend to be longer~\citep{zeng2025simplerlzooinvestigatingtamingzero},
% We also discuss a variant of \mtwo{}, \mthree{}, which explicitly encourages additional exploration on incorrect responses, allowing models to discover potential correct patterns through extended reasoning.

We conduct comprehensive experiments on three reasoning models ranging from 1.5B to 32B parameters, across four challenging benchmarks: MATH500, AIME2024, AMC2023, and OlympiadBench. Our extensive evaluations demonstrate that our proposed \mone{} series outperform existing works, while \mtwo{} and its variant \mthree{} achieve the best Pareto-optimal balance between accuracy and token efficiency. Unlike methods that improve token efficiency at the expense of accuracy, our proposed approaches deliver substantial gains in both dimensions—even on the challenging AIME2024 benchmark. For example, applying \mtwo{}/\mthree{} to \dss{} improves accuracy by $+\bf{6.1}$ percentage points while reducing token usage by $\bf{63}$\% on AIME24. Our further analysis reveals that after these RL-based CoT compressions, the reasoning behaviors of LRMs become more concise and demonstrate improved quality with fewer redundant and unhelpful ``self-reflection''.
% \yz{other than the numbers, do we have some scientific observation, insights we can talk about?}
% Experiments on with significant token efficiency and performance, further demonstrate the effectiveness of our proposed methods.\jh{we need to describe the specific number, like how many tokens are reduced}

% Our experiments on Deepseek-R1-Distilled-1.5B \wl{TODO: extensive runs?} demonstrates that compared to previous methods, truncation method and RTL method, DARL and its variant achieve best pareto optimal between accuracy and token efficiency. Rather than increasing token efficiency by compromising accuracy, RTL and DARL achieve both significant token efficiency and accuracy boost, even in the AIME24 benchmark. Experimetns on \wl{TORUN} DeepScaleR-1.5B-Preview, Deepseek-R1-Distilled-7B with significant tokens effieicny and performance further demosntrate the effectivess of our proposed method.

% In a nutshell, our contributions can be concluded as 1. truncation 2. unified reward 3. difficulty-aware adaptive rewards.

\section{Preliminary}

\paragraph{Enhancing Reasoning via RL}

RL has been demonstrated as an effective way to train strong large reasoning models~\citep{deepseekai2025deepseekr1incentivizingreasoningcapability,openai2024openaio1card,kimiteam2025kimik15scalingreinforcement} across different domains like math~\citep{zeng2025simplerlzooinvestigatingtamingzero}, coding~\citep{deepseekai2025deepseekr1incentivizingreasoningcapability} and agentic tasks~\citep{openai2024openaio1card}. For example, using a simple rule-based outcome reward~\citep{deepseekai2025deepseekr1incentivizingreasoningcapability,zeng2025simplerlzooinvestigatingtamingzero}, the mathematical reasoning capabilities of models can be substantially improved after RL training, often accompanied by the emergence of “self-reflection” style thinking patterns. Following these previous works, we leverage rule-based reward designed as a simple scoring system~\citep{deepseekai2025deepseekr1incentivizingreasoningcapability}: +1 for correct responses, -0.5 for incorrect responses, and -1 for responses with invalid format.

Suppose \( x \) is the question and \( y \) is the response generated by the models, the optimization objective with KL-constrained~\citep{schulman2017proximal} in RL can be formulated as:

\begin{equation}
    \pi_{\theta}^* = \operatorname*{\arg\max}_{\theta} \mathbb{E}_{x \sim \mathcal{D}} \left[ \mathbb{E}_{y \sim \pi(\cdot|x)} [R(x, y)] - \beta \mathbb{D}_{KL}[\pi_{\theta}(\cdot|x) \,||\, \pi_{ref}(\cdot|x)] \right]
    \label{eq:rl}
\end{equation}

where \( R(x, y) \) represents the reward of the entire trajectory, and \(\pi_{ref}\) is the reference model, which is the model prior to the RL training phase. $\beta$ is the parameter to control the two optimization targets. In this paper, we utilize GRPO~\citep{shao2024deepseekmath} to optimize this objective.

\paragraph{RL for Efficient Reasoning}  
In addition to enhancing reasoning capabilities, RL also holds promise for improving token efficiency in LRMs~\citep{kimiteam2025kimik15scalingreinforcement,qu2025surveyefficientreasoninglarge}. Several approaches have been proposed to this end. Most methods involve reward shaping, where models are incentivized to produce shorter responses by associating higher rewards with more concise outputs~\citep{kimiteam2025kimik15scalingreinforcement,aggarwal2025l1controllinglongreasoning,arora2025traininglanguagemodelsreason}.

% \jh{to compress space, I think we can just talk RL for efficient reasoning as a separate paragraph rather than as a section. And no need to have subsection 2.1 and 2.2 here.}

% The objective in Eq.~\ref{eq:rl} can be further formulated as:\jh{This constrained optimization formulation is kinda weird to me, is any previous length-based RL works formulating this way?}

% \begin{align}
%     \pi_{\theta}^* = \operatorname*{\arg\max}_{\theta} \, \mathbb{E}_{x \sim \mathcal{D}} \left[ \mathbb{E}_{y \sim \pi(\cdot|x)} [R(x, y)] 
%     - \beta \mathbb{D}_{KL}[\pi_{\theta}(\cdot|x) \,||\, \pi_{ref}(\cdot|x)] \right] & \nonumber \\
%      \text{subject to } \mathbb{E}_{x \sim \mathcal{D}, y \sim \pi_{\theta}(\cdot|x)}(len(y)) \leq \tau &
% \end{align}

% The constraint can be implemented using various approaches. 

% \section{Truncation: \jhc{A Simple Approach}{A Simple Yet Effective Baseline}}
\section{Truncation: A Simple Yet Effective Baseline}
\label{sec:truncation}

In this section, we start from a simple yet effective baseline, where we simply set max generation length to a value significantly smaller than the model's original context window during RL training—for example, 8,192 tokens versus 32,768 in DeepSeek-R1-Distilled models.
Intuitively, this approach truncates the responses beyond the context window and regards them as incorrect, thus it
pushes the model to generate accurate yet more concise responses under strict token constraints. This baseline has been explored recently in concurrent works~\cite{hou2025thinkprunepruninglongchainofthought,deepscaler2025}.
% \jh{plz note that in this change, also following the storyline of intro, we don't present this baseline as we revisit xxx found by others before -- it is not the actual logic of our project, actually we independently found this early, thus we present it directly, and cite related works as concurrent works}
In our experiments, we adopt \dss{} as the base model and investigate the effects of truncation by limiting maximum generation lengths to 4,096, 6,144, and 8,192 tokens.

\begin{table}[t!]
    \centering
    % \caption{\jh{caption needs to be self-complete, for example, here at least it needs to mention these are results of the truncation baseline}Accuracy (\%) with average token usage for each dataset. ``Original'' denotes the original \dss. $\operatorname{T}_{k}$ denotes the models after RL training with context window $k$.}
    \caption{Results of baseline truncation method with different context window. $\operatorname{T}_{k}$ denotes the models after RL training with context window $k$. Accuracy (\%) with average token usage for each dataset. ``Original'' denotes the original \dss.}  
    \resizebox{\textwidth}{!}{%
      \begin{NiceTabular}{l|ccccc|ccccc}
        \toprule
        \midrule
          & \multicolumn{5}{c}{\textbf{Accuracy (\%)}}
          & \multicolumn{5}{c}{\textbf{Generation Length (tokens)}} \\
          & \makecell{MATH\\500} & AIME  & AMC   & \makecell{Olympiad\\Bench} & Avg.
          & \makecell{MATH\\500} & AIME  & AMC   & \makecell{Olympiad\\Bench} & Avg. \\
        \midrule
        Original
          & 83.9 & 28.9 & 71.6 & 43.3 & 56.9
          & 5042 & 15956 & 8202 & 11510 & 10177 \\
        $T_{8192}$
          & 81.8 & 24.8 & 70.9 & 43.9 & 55.3
          & 1795 & 4465 & 2560 & 2841  & 2915  \\
        $T_{6144}$
          & 80.9 & 20.2 & 66.2 & 42.1 & 52.3
          & 1351 & 2821 & 1917 & 1947  & 2009  \\
        $T_{4096}$
          & 77.7 & 19.2 & 62.2 & 38.5 & 49.4
          & 1054 & 2481 & 1484 & 1564  & 1646  \\
        \midrule
        \bottomrule
      \end{NiceTabular}%
    }
    \label{tab:truncation}
    % \vspace{-10pt}
  \end{table}

% \begin{table}[t!]
% \centering
% % \renewcommand{\arraystretch}{1.3}
% \caption{Accuracy (\%) with average token usage as blue subscript for each dataset. ``Original'' denotes the original \dss. $\operatorname{T}_{k}$ denotes the models after RL training with context window $k$.}
% \begin{tabular}{lccccc}
% \toprule
% \textbf{Model} & \textbf{MATH500} & \textbf{AIME} & \textbf{AMC} & \textbf{OlympiadBench} & \textbf{Average} \\
% \midrule
% Original & 
% 83.9\textsubscript{\textcolor{blue}{5042}} & 
% 28.9\textsubscript{\textcolor{blue}{15956}} & 
% 71.6\textsubscript{\textcolor{blue}{8202}} & 
% 43.3\textsubscript{\textcolor{blue}{11510}} & 
% 56.9\textsubscript{\textcolor{blue}{10177}} \\
% $\operatorname{T}_{8192}$ & 
% 81.8\textsubscript{\textcolor{blue}{1795}} & 
% 24.8\textsubscript{\textcolor{blue}{4465}} & 
% 70.9\textsubscript{\textcolor{blue}{2560}} & 
% 43.9\textsubscript{\textcolor{blue}{2841}} & 
% 55.3\textsubscript{\textcolor{blue}{2915}} \\
% $\operatorname{T}_{6144}$ & 
% 80.9\textsubscript{\textcolor{blue}{1351}} & 
% 20.2\textsubscript{\textcolor{blue}{2821}} & 
% 66.2\textsubscript{\textcolor{blue}{1917}} & 
% 42.1\textsubscript{\textcolor{blue}{1947}} & 
% 52.3\textsubscript{\textcolor{blue}{2009}} \\
% $\operatorname{T}_{4096}$ & 
% 77.7\textsubscript{\textcolor{blue}{1054}} & 
% 19.2\textsubscript{\textcolor{blue}{2481}} & 
% 62.2\textsubscript{\textcolor{blue}{1484}} & 
% 38.5\textsubscript{\textcolor{blue}{1564}} & 
% 49.4\textsubscript{\textcolor{blue}{1646}} \\
% \bottomrule
% \end{tabular}
% \label{tab:truncation}
% \end{table}

\paragraph{Effectiveness of Truncation} Table~\ref{tab:truncation} presents the performance of models across various benchmarks under different truncation sizes. Compared to the original model, surprisingly, RL training with a context window of 8192 tokens achieves a substantial $\mathbf{71\%}$ improvement in token efficiency, while maintaining competitive accuracy with a 1.6 absolute point degradation on average. This demonstrates that truncation is a simple yet effective approach for enhancing reasoning efficiency in LRMs.

\paragraph{Efficacy-Efficiency Trade-off} Although truncation proves effective on average across benchmarks, its impact varies significantly with task difficulty. A closer look at the results reveals that performance on the most challenging benchmark, AIME, suffers a notable $\mathbf{4.1}$ drop in accuracy under the 8192 token limit. When the context window is further reduced to 4096, the accuracy on AIME deteriorates even more sharply, with a $\mathbf{9.7}$ decline, by far the largest drop observed, compared to only a $\mathbf{7\%}$ decrease on MATH500. This highlights that the benefits of truncation involve a trade-off: while it improves efficiency overall, it may disproportionately affect harder tasks. To better understand this disproportionate performance drop on harder benchmarks, we note that the truncation ratio during training is initially very high (Figure~\ref{fig:clipratio1}), exceeding $\bf{45}\%$, and remains above $\bf{10}\%$ even after 200 rollout steps. Specifically, for the AIME dataset, over $\bf{75}\%$ of responses exceed 8192 tokens, compared to only $\bf{15}\%$ for MATH500. This indicates that truncation disproportionately impacts more complex tasks like AIME, where long reasoning trajectories are often necessary. Next, we formulate the truncation baseline from the reward shaping perspective, and connect it with related works.

\section{A Unified View on Efficient Reasoning with RL}
\label{sec:unified}
% \jh{this introduction looks unrelevant to Section 3 at all. i tried a better transition}
In this section, we aim to understand the truncation baseline and other RL-based efficient reasoning approaches through a unified perspective. We first connect them together via length-based reward shaping, and then derive new alternatives with this view.
% \jhc{In this section, we begin by unifying various reinforcement learning approaches for efficient reasoning. We observe that most current long-to-short RL methods can be integrated into a framework focused on reward design based on sequence length. First, we derive an equivalent form to unify different long-to-short RL methods including the truncation method. Subsequently, we propose a novel reward design inspired by this unified form and the truncation method.}{}

\subsection{The Unified Formulation}
Here we first present a unified formulation, and then we show how the truncation baseline and other works fit into this formulation.
Specifically, we define the reward function with two parts:
a correctness term \( C(y) \) and a length-based term \( S(y) \) controlled by a control variable $\lambda(y)$: 

\begin{equation}
    \hat{R}(x, y) = C(y) + \lambda(y)\cdot S(y)
    \label{eq:unified_reward}
\end{equation}

In most length reward methods, \( C(y) = R(x,y) \), representing the original rule-based reward for correctness. However, in truncation-based approaches, \( C(y) = 0 \) as we discuss below. The term \( S(y) \) denotes the length reward, which varies across different methods. 

\paragraph{Formulating the Truncation Baseline}
As shown in Table~\ref{tab:unified_reward}, truncation is a special case of the length reward with \(\csymbol{}=0\), where the target length \(L_T\) is enforced by the context window. ThinkPrune~\citep{hou2025thinkprunepruninglongchainofthought} is another truncation-based approach, which extends vanilla truncation by introducing an adaptive target length $L_A$ to replace fixed target length $L_T$. They iteratively choose $L_A$ and separate their training into three stages. Table~\ref{tab:unified_reward} also outlines other formulations, we will introduce them individually in the following sections.
\subsection{Connecting Previous Efforts Together}
% \jh{in this part, we build on our unified formulation in eq.xxx and aim to connect previous approaches, below we describe several main catogries of them..}
% designs for \( \csymbol{} \), \( \lambda(y) \) and \( S(y) \).
In this part, we build on our unified formulation in Eq.~\ref{eq:unified_reward} and aim to connect previous approaches, below we describe several main categories of them. Table~\ref{tab:unified_reward} formulates different length-based reward shaping by different designs for \( \csymbol{} \), \( \lambda(y) \) and \( S(y) \). Parameter \( \alpha \) is a hyperparameter coefficient that controls the magnitude of the length reward \( S(y) \). We provide detailed explanations for each formulation in Appendix~\ref{ap:details_unified}.

% Besides the truncation method, integrating length rewards into the RL training process is a widely used approach for enhancing reasoning efficiency~\citep{kimiteam2025kimik15scalingreinforcement,aggarwal2025l1controllinglongreasoning,arora2025traininglanguagemodelsreason}. These length rewards can be categorized based on their objectives. For example, Kimi-k1.5~\citep{kimiteam2025kimik15scalingreinforcement} and Efficient Reasoning~\citep{arora2025traininglanguagemodelsreason} employ a group-based reward, where shorter responses receive higher rewards compared to other responses for the same question within a rollout group. Alternatively, a budget-based approach, such as L1~\citep{aggarwal2025l1controllinglongreasoning}, sets an exact or maximum budget. Responses that deviate from or exceed this budget are penalized more severely based on the degree of divergence. We present a unified paradigm that encompasses these different length rewards. This unified paradigm not only integrates various length rewards but also incorporates the truncation method, serving as a bridge to connect length rewards and truncation methods.

% \paragraph{The Unified Reward}

\begin{table}[t!]
    \centering
    \caption{
    Formulation of different approaches based on Eq.~\ref{eq:unified_reward}. $\csymbol{}$ is mainly for correctness, $S(y)$ is the length reward, and $\lambda(y)$ is a control variable to control how length reward is applied. $\mathbb{I}(R)$ stands for $\mathbb{I}(R(x,y)=1)$ and $\mathbb{I}(\cdot)$ is an indicator function. $\rho$ is the negative reward given for incorrect responses. $L(y)$ is the length of the generated response. \( \alpha \) is the coefficient that controls the magnitude of the length reward. The shapes of different rewards are shown in the visualization, where x axis is the length of the response.\textcolor[rgb]{0.28, 0.39, 0.70}{Blue} represents the curve for correct responses, while \textcolor[rgb]{0.66, 0.24, 0.19}{Red} represents the curve for incorrect responses. For approaches, ThinkPrune, \mtwo{} and \mthree, there are different lines with similar colors indicate that the reward is dynamic which is realized by different $L_A$ values. The details of visualization are available in the Appendix~\ref{ap:vis}.}
    \renewcommand{\arraystretch}{1.5}  % 设置一个适中的行高
    \resizebox{\textwidth}{!}{%
    \begin{tabular}{
        l
        >{\centering\arraybackslash}p{5cm}
        >{\centering\arraybackslash}p{6cm}
        >{\centering\arraybackslash}m{3cm}  % ← 将最后一列改为 m 列格式，垂直居中
    }
    \toprule
    \textbf{Name} & $\csymbol{}, \lambda(y)$ & $S(y)$ & \textbf{Visualization} \\
    \midrule
    \multicolumn{4}{l}{\textbf{Truncation Method}} \\
    \cmidrule[0.3pt](l{0pt}r{0pt}){1-4}
    Vanilla Truncation 
        & $0, 1$ 
        % & $0, R(x, y)$ 
        % & $\mathbb{I}(L(y) \leq L_T) + \frac{\rho}{R(x,y)} \cdot \mathbb{I}(L(y) > L_T)$ 
        & $\begin{cases}
            R(x,y) & \text{if } L(y) \leq L_T \\
            \rho & \text{if } L(y) > L_T
          \end{cases}$
        & \includegraphics[width=1.4cm]{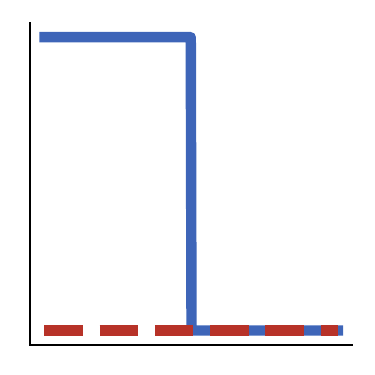} \\
    \cdashline{1-4}[0.5pt/2pt]
    ThinkPrune~\citep{hou2025thinkprunepruninglongchainofthought} 
        & $0, 1$ 
        &
        $
        \begin{cases}
            R(x,y) & \text{if } L(y) \leq L_A \\
            \rho & \text{if } L(y) > L_A
          \end{cases}
        $
        % & $\mathbb{I}(L(y) \leq L_A) + \frac{\rho}{R(x,y)} \cdot \mathbb{I}(L(y) > L_A)$ 
        & \includegraphics[width=1.4cm]{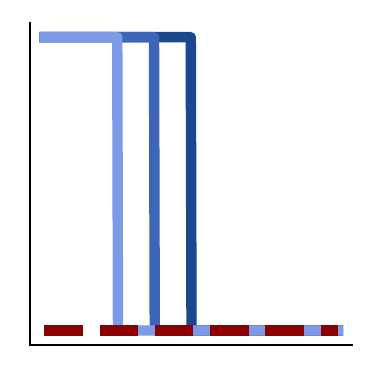} \\
    \cmidrule[0.3pt](l{0pt}r{0pt}){1-4}
    \multicolumn{4}{l}{\textbf{Group-based Reward}} \\
    \cmidrule[0.3pt](l{0pt}r{0pt}){1-4}
    Efficient Reasoning~\citep{arora2025traininglanguagemodelsreason} 
        & $R(x,y), \mathbb{I}(R)$
        & $-\alpha \cdot \sigma\!\Bigl(\frac{L(y)-\mathrm{Mean}(y)}{\mathrm{STD}(L)}\Bigr)$ 
        & \includegraphics[width=1.4cm]{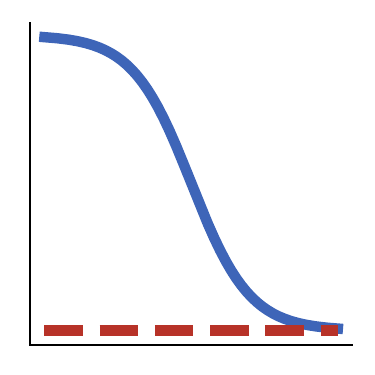} \\
    \cdashline{1-4}[0.5pt/2pt]
    Kimi-k1.5~\citep{kimiteam2025kimik15scalingreinforcement} 
        & $R(x,y), 1$
        &
        $\displaystyle
        \begin{cases}
          0.5-\tfrac{L(y)-L_{\min}}{L_{\max}-L_{\min}} & \text{if } \mathbb{I}(R) = 1 \\
          \min\!\left(0,\;0.5-\tfrac{L(y)-L_{\min}}{L_{\max}-L_{\min}}\right) & \text{if } \mathbb{I}(R) = 0
        \end{cases}
        $
        % & $\displaystyle
        %     \mathbb{I}(R)\cdot\Bigl(0.5-\tfrac{L(y)-L_{\min}}{L_{\max}-L_{\min}}\Bigr)
        %   + (1-\mathbb{I}(R))\cdot\min\!\Bigl(0,\;0.5-\tfrac{L(y)-L_{\min}}{L_{\max}-L_{\min}}\Bigr)$
        & \includegraphics[width=1.4cm]{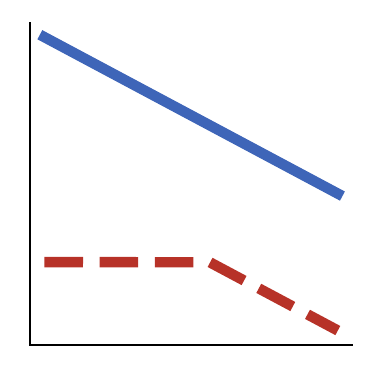} \\
    % \addlinespace
    \cmidrule[0.3pt](l{0pt}r{0pt}){1-4}
    \multicolumn{4}{l}{\textbf{Budget-based Reward}} \\
    \cmidrule[0.3pt](l{0pt}r{0pt}){1-4}
    L1-Exact~\citep{aggarwal2025l1controllinglongreasoning} 
        & $R(x,y), 1$ 
        & $-\alpha\cdot |L(y) - L_T|$ 
        & \includegraphics[width=1.4cm]{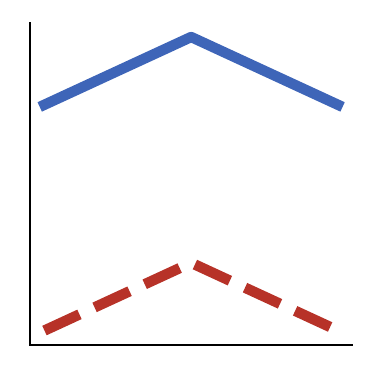} \\
    \cdashline{1-4}[0.5pt/2pt]
    L1-Max~\citep{aggarwal2025l1controllinglongreasoning} 
        & $0, \mathbb{I}(R)$ 
        & $ \operatorname{clip}(\alpha \cdot (L(y) - L_T) + \delta, 0, 1)$ 
        & \includegraphics[width=1.4cm]{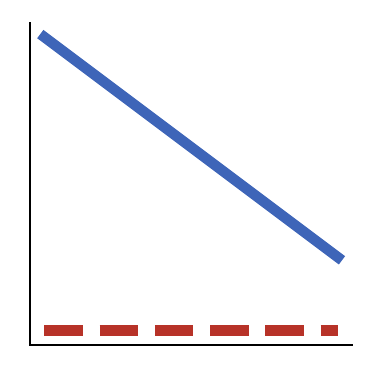} \\
    \cmidrule[0.3pt](l{0pt}r{0pt}){1-4}
    \multicolumn{4}{l}{\textbf{Length-Based Step Reward and Variants}} \\
    \cmidrule[0.3pt](l{0pt}r{0pt}){1-4}
    \mone 
        & $R(x,y), \mathbb{I}(R)$ 
        & $\alpha \cdot \mathbb{I}(L(y) \leq L_T)$ 
        & \includegraphics[width=1.4cm]{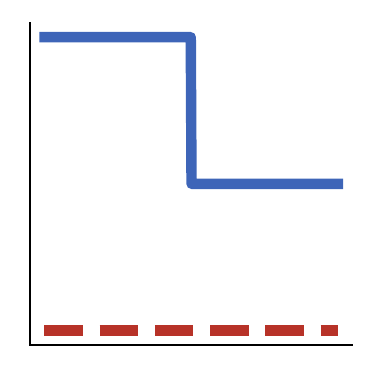} \\
    \cdashline{1-4}[0.5pt/2pt]
    \mtwo 
        & $R(x,y), \mathbb{I}(R)$ 
        & $\alpha \cdot \mathbb{I}(L(y) \leq L_A)$ 
        & \includegraphics[width=1.4cm]{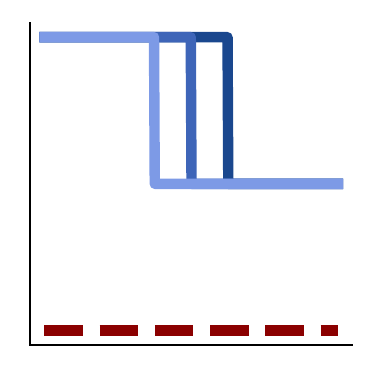} \\
    \cdashline{1-4}[0.5pt/2pt]
    \mthree 
        & $R(x,y), 1$ 
        &  
        % $\displaystyle
        % \begin{cases}
        %   \alpha & \text{if } \mathbb{I}(R)=1 \text{ and } L(y)\leq L_A \\
        %   \alpha & \text{if } \mathbb{I}(R)=0 \text{ and } L(y) > L_A \\
        %   0 & \text{otherwise}
        % \end{cases}$
        $\displaystyle
            \alpha\cdot\mathbb{I}(R)\cdot\mathbb{I}(L(y)\leq L_A)
          + \alpha\cdot(1-\mathbb{I}(R))\cdot\mathbb{I}(L(y)> L_A)$
        & \includegraphics[width=1.4cm]{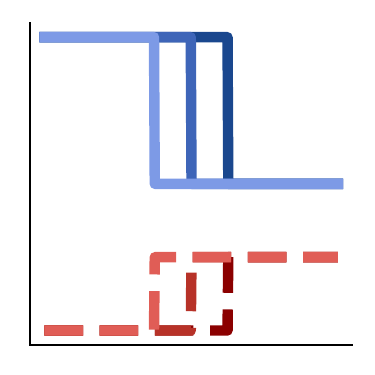} \\
    \bottomrule
    \end{tabular}%
    }
    \vspace{-0.6cm}
    \label{tab:unified_reward}
\end{table}

\paragraph{Group-based Reward} In group-based reward, the length reward $S(y)$ is designed to encourage brevity by assigning higher scores to shorter responses within a rollout group, such as Efficient Reasoning~\citep{arora2025traininglanguagemodelsreason} and Kimi-k1.5~\citep{kimiteam2025kimik15scalingreinforcement} as formulated in Table~\ref{tab:unified_reward}. However, this comparison-based approach can lead to reward hacking. Models tend to exploit $S(y)$ by generating overly concise responses, particularly for simpler questions. We demonstrate this phenomenon for the Efficient Reasoning baseline in Figure~\ref{fig:training_accuracy} and Figure~\ref{fig:reward_comparison} in Appendix~\ref{ap:acc_rewards}, where training accuracy initially decreases while total reward increases. Additionally, Table~\ref{tab:main_results} shows a more significant drop in MATH500 accuracy compared to other methods, further supporting this observation.

\paragraph{Budget-based Reward} 
Budget-based rewards use query-specific target lengths (budgets) and penalize responses that deviate from these instructions. While this mitigates reward hacking seen in group-based schemes, it can destabilize training. Models require exposure to diverse budgets, but in large context windows (e.g., 16,384 tokens), these targets become sparsely distributed, causing reward fluctuations. Figure~\ref{fig:reward_comparison} illustrates this instability. Using L1-Max as a representative method, we observe that with smaller contexts (4,096 tokens), it achieves stable reward increases comparable to other methods. However, with 16,384-token contexts, rewards fluctuate significantly and underperform alternative approaches.

\subsection{Bridging the Gap: Length-based Step Reward}
\label{sec:lsr}

% \jh{this section is too long. Generally just first talk about the issue of truncation baseline, say it just simply truncation, may over-penalize. So we just directly extend it and introduce the target length, and that is it.}

% \jh{according to the current logic of intro, we should discuss the results of \mone{} here, just briefly. For example, we may just discuss figure 1.}

% As shown in Eq~\ref{eq:unified_reward} and visualized in Table~\ref{tab:unified_reward}, a key issue with the truncation method is that it omits the \(\csymbol{}\) term and reduces the reward to a multiplicative form \(\lambda(y)\cdot S(y)\), conflating length and accuracy. There is an area where the rewards for correct and incorrect responses are mixed (the overlap between blue and red lines). Models cannot learn from the accuracy reward in that area. Consequently, the model receives no learning signal from correct responses that exceed the target length \(L_T\). This issue is magnified in more challenging questions, as they tend to require longer responses. This can also be observed in Table~\ref{tab:main_results}.
As shown in Eq.~\ref{eq:unified_reward} and visualized in Table~\ref{tab:unified_reward}, a key limitation of the truncation method is that it assigns the same penalties to overlong responses as it does to incorrect ones, which may over-penalize long but correct explorations. To address this issue, we extend it as a novel reward shaping approach called \textbf{L}ength-b\textbf{A}sed \textbf{S}t\textbf{E}p \textbf{R}eward (\mone), which adopts a step reward function guided by a target length, rather than performing hard truncation.

Specifically, we design the length reward term $S(y)$ as an indicator function based on a target length $L_T$. This function assigns a length-based bonus to responses shorter than $L_T$. We also set the context window significantly larger than the target length $L_T$ (e.g., 16,384 vs. 4,096) where truncation rarely happens. And the length reward term $S(y)$ is only activated when responses are correct, thereby improving the efficacy-efficiency trade-off. As visualized in Table~\ref{tab:unified_reward}, \mone{} closely resembles the vanilla truncation approach; the only difference is that, instead of truncating long responses, it awards bonus rewards to correct responses that do not exceed the target length. To balance the correctness reward $C(y)$ and length reward $S(y)$, we follow a typical setting and set $\alpha$ as $0.5$.

Empirical results are demonstrated in Figure~\ref{fig:pareto} and Table~\ref{tab:main_results}, training with the \mone{} reward achieves improved Pareto-optimality compared to all previous methods. Notably, it is the first approach to simultaneously deliver significant improvements in both accuracy and token efficiency on the challenging AIME24 benchmark. These results establish \mone{} as a promising reward design framework for enhancing the balance between efficacy and efficiency.

\section{Adaptive Length-based Step Reward Shaping}
\label{sec:lsrd_overview}

\subsection{Design Principles}
We highlight two key limitations not addressed in the design of \mone{}: (1) \mone{} requires specifying a fixed target length prior to training; however, as the model evolves during training, the optimal response length may also change and should ideally adapt dynamically. (2) Additionally, different questions  demand reasoning traces of varying lengths—simple questions may be effectively addressed with shorter reasoning, while more complex questions benefit from longer, more detailed deliberation.

Therefore, we extend \mone{} to be \textbf{D}ynamic and \textbf{D}ifficulty-aware, which we term as \mtwo{}. Rather than using a single fixed target length, our approach dynamically adjusts the target length throughout training and tailors it to questions of varying difficulty. 
Concretely, \emph{\mtwo{} decouples the target length hyperparameter across different queries, allowing distinct target lengths to be assigned to various queries. Moreover, these target length hyperparameters are dynamically adjusted throughout training.} 

We separate queries into three buckets of easy, medium, and hard difficulty levels, based on the correctness rates within the rollout batch -- for each question, we have $k$ rollouts and use thresholds $k/3$ and $2k/3$ to separate them. As such, we have three distinct target length hyperparameters for these three query groups.
Notably, we perform difficulty assessment for the queries during real-time RL training and use the training rollout batch, thus it only incurs negligible overhead on the computation.
Being dynamic and difficulty-aware, one challenge raised is how to set the dynamic processes of the decoupled target length hyperparameters. Next, we introduce an \emph{automatic} adapting mechanism, to adapt them without any manual intervention.

\subsection{Automatic Adapting Mechanism}
\label{sec:adaptauto}
% \paragraph{Difficulty Classification} \mtwo{} classifies questions into three difficulty levels—easy, medium, and hard—based on their correctness ratio. For each question, we have $K$ rollouts, so we can use thresholds of $\frac{K}{3}$ and $\frac{2K}{3}$ as pivots to separate these three levels. This classification applies consistently during both monitoring and training phases.
% \jh{i suggest to call it adapting rather than monitoring, plz change other places of the paper}
% \paragraph{Monitoring Mechanism} 
\mtwo{} is driven by an automatic adapting mechanism that periodically evaluates and adjusts the target length parameters ($L_A$ in Table~\ref{tab:unified_reward}) for each difficulty level. 
Specifically, we first extract a small monitoring dataset $\mathcal{D}^{M}$ (e.g., 500 samples) from training data that mirrors the distribution of the training data. Every N training steps (e.g., 20), our approach searches and sets the target length hyperparameters based on this monitoring dataset. 

% For each question in the monitoring dataset, K responses are generated using the current policy model. Then for each difficulty level $d$, the adapting mechanism determines appropriate scaling factors $\beta_d$, which are used to adjust the effective target length ($\beta_d \cdot L_T$) as shown in Table~\ref{tab:unified_reward}.
% Intuitive
Denote the three-class difficulty level of a query as $d$, to determine the target length hyperparameters, we propose a metric called \textbf{E}xpected \textbf{C}orrect \textbf{R}esponses (ECR), which estimates how many complete, correct responses we can expect for each difficulty level given response length limits. Formally, we sample $K$ responses for each query in the monitoring set,\footnote{Practically, $K$ is set to be the same as the rollout size used during training, in order to maintain consistency with the training scenario.} and ECR is computed as

\begin{equation}
    ECR_{d} = P_{l,d} \cdot |C_d|
\end{equation}

where $P_{l,d}$ is the coverage ratio (proportion of responses that fit within a given token length $l$). The value $|C_d|$ is fixed for each difficulty group. Since we use the ratio of correct responses within each rollout group to determine the difficulty level, there is a minimum number of correct responses for each group (e.g., 6, 3, and 1 correct responses for easy, medium, and hard levels, respectively, when $K=8$). We set $|C_d|$ as these minimum values for each group.

The monitoring module enumerates potential target lengths from the lower bound target length $L_T$ tokens up to the maximum context window (16,384 tokens) with an interval of $I$, computing coverage ratios $P_{l,d}$ for each length. We select the smallest target length as the adaptive target length $L_A$ satisfying $ECR_d \geq 1$ for each difficulty level $d$, ensuring at least one complete and correct response. 

Intuitively, this mechanism sets the target length as the minimal generation length such that at least one rollout response is expected to be correct. This approach is reasonable because generating sequences shorter than this length would likely be detrimental, as correct responses are less probable. Conversely, generating longer sequences may be redundant, since correct responses can already be obtained with a shorter generation length.
% The scaling factor $\beta_d$ is then determined such that $\beta_d \cdot L_T$ equals the selected target length.

\paragraph{Dynamic and Difficulty-Aware Reward} During training, we apply these monitoring-derived parameters to implement dynamic and difficulty-aware rewards. Each training question's difficulty level is determined using the same classification method described earlier. Easier questions receive smaller target lengths (i.e. smaller scaling factor $\beta$), while harder questions receive larger ones (i.e. larger scaling factor $\beta$). Since monitoring runs every N steps, the difficulty-dependent target lengths are automatically adapted to the evolving policy model.

\paragraph{Computational Efficiency} This automatic adapting mechanism adds minimal computational overhead. By using a small monitoring dataset and evaluating only periodically, our method increases computation by just 3.5\% in our experiments.

\subsection{\mthree: A Variant of \mtwo{} to Encourage Exploration}
\label{sec:lsrde}

Previous works~\citep{muennighoff2025s1simpletesttimescaling,hou2025thinkprunepruninglongchainofthought} find that with more test-time compute, the reasoning ability of models will improve. Meanwhile, some other works~\citep{zeng2025simplerlzooinvestigatingtamingzero} show that incorrect responses tend to produce more tokens. Both findings are related to the exploration of policy models, where models try to explore by consuming more compute to get the correct answers. Therefore, we further propose a variant of \mtwo, named \mthree, to encourage the exploration of policy models for those incorrect responses. The only difference for \mthree{} is to encourage those incorrect responses to be further explored to find correct pattern by applying reduced penalties to those that are incorrect and exceed the target length. The form of \mthree{} can be seen in Table~\ref{tab:unified_reward}.

% Previous works~\citep{muennighoff2025s1simpletesttimescaling,hou2025thinkprunepruninglongchainofthought} find that with more test-time compute, the reasoning ability of models will improve. Meanwhile, some other works~\citep{zeng2025simplerlzooinvestigatingtamingzero} show that incorrect responses tend to produce more tokens. Both findings are related to the exploration of policy models, where models try to explore by consuming more compute to get the correct answers. Therefore, we further propose a variant of \mtwo, named \mthree, to encourage the exploration of policy models for those incorrect responses. The key difference in \mthree{} is that it promotes further exploration of incorrect responses The form of \mthree{} can be seen in Table~\ref{tab:unified_reward}.

\section{Experiments}
\label{sec:exp}

\begin{table}[t!]
    \centering
    \caption{Accuracy (\%) with average token usage for each dataset and different methods. Most important results in this table are visualized in Figure~\ref{fig:pareto} and Figure~\ref{fig:pareto_full} in Appendix~\ref{ap:full_pareto_figures}. The base model is \dss{}. "Original" denotes the original model. $T_k$ is the truncation method with context window $k$. ``Group'' denotes the Efficient Reasoning~\citep{arora2025traininglanguagemodelsreason} with different $\alpha$. Due to the space limit, we only show three most representative results here. For the full results, please refer to Tabel~\ref{tab:main_results_full} in Appendix~\ref{ap:full_results}.}
    \resizebox{\textwidth}{!}{%
    \begin{NiceTabular}{lccccc|ccccc}[code-before=\rowcolors{17-25}{lightblue}{lightblue}]
        \toprule
        \midrule
          & \multicolumn{5}{c}{\textbf{Accuracy (\%)}}
          & \multicolumn{5}{c}{\textbf{Generation Length (tokens)}} \\
          & \makecell{MATH\\500} & AIME  & AMC   & \makecell{Olympiad\\Bench} & Avg.
          & \makecell{MATH\\500} & AIME  & AMC   & \makecell{Olympiad\\Bench} & Avg. \\
        \midrule
        % \multicolumn{11}{c}{\dss{}} \\
        % \midrule
        Original
          & 83.9 & 28.9 & 71.6 & 43.3 & 56.9
          & 5042 & 15956 & 8202 & 11510 & 10177 \\
        $T_{8192}$
          & 81.8 & 24.8 & 70.9 & 43.9 & 55.3
          & 1795 & 4465 & 2560 & 2841  & 2915  \\
        $T_{6144}$
          & 80.9 & 20.2 & 66.2 & 42.1 & 52.3
          & 1351 & 2821 & 1917 & 1947  & 2009  \\
        $T_{4096}$
          & 77.7 & 19.2 & 62.2 & 38.5 & 49.4
          & 1054 & 2481 & 1484 & 1564  & 1646  \\
        $\text{Group}_{\alpha=0.4}$
          & 74.6 & 25.0 & 69.2 & 43.1 & 53.0
          & 1069 & 4747 & 2162 & 2536 & 2629 \\
       $\text{Group}_{\alpha=0.2}$
          & 78.1 & 28.1 & 68.0 & 44.4 & 54.7
          & 1135 & 5628 & 2635 & 2944 & 3085 \\
       $\text{Group}_{\alpha=0.1}$
          & 77.0 & 29.0 & 69.5 & 44.9 & 55.1
          & 1228 & 6301 & 2808 & 3271 & 3402 \\
        $\text{Group}_{\alpha=0.05}$
          & 74.4 & 30.2 & 65.5 & 43.1 & 53.3
          & 1193 & 4839 & 2457 & 2703 & 2798 \\
        L1-Max-1024
          & 76.4 & 15.0 & 59.4 & 39.1 & 47.5
          & 661 & 1303 & 933 & 938 & 959 \\
        L1-Max-4096
          & 79.7 & 20.0 & 65.0 & 41.0 & 51.4
          & 875 & 1718 & 1159 & 1229 & 1245 \\
        $\text{\mone}_{L_T=2048}$
          & 83.6 & 29.2 & 71.6 & 44.1 & 57.1
          & 1913 & 4815 & 2493 & 2767 & 2895 \\
         $\text{\mone}_{L_T=4096}$
          & 83.9 & 31.0 & 74.1 & 45.7 & 58.7
          & 1914 & 5915 & 3136 & 3579 & 3636 \\
         $\text{\mone}_{L_T=8192}$
          & 85.6 & 31.5 & 75.9 & 47.7 & 60.2
          & 2736 & 6589 & 4162 & 4547 & 4509 \\
         $\text{\mtwo}_{L_T=1024}$
          & 83.0 & 30.6 & 72.8 & 43.7 & 57.5
          & 1362 & 4991 & 256 & 2837 & 2862 \\
         $\text{\mtwo}_{L_T=2048}$
          & 82.2 & 31.0 & 73.3 & 46.2 & 58.2
          & 1623 & 5158 & 2572 & 2960 & 3059 \\
         $\text{\mtwo}_{L_T=4096}$
          & 84.2 & 34.2 & 75.3 & 47.3 & 60.3
          & 1872 & 5750 & 2981 & 3474 & 3520 \\
         $\text{\mthree}_{L_T=1024}$
          & 82.1 & 33.8 & 72.2 & 43.7 & 58.0
          & 1350 & 4794 & 2254 & 2654 & 2763 \\
         $\text{\mthree}_{L_T=2048}$
          & 83.9 & 31.5 & 75.3 & 46.4 & 59.3
          & 1456 & 5263 & 2679 & 2971 & 3092 \\
         $\text{\mthree}_{L_T=4096}$
          & 83.5 & 35.0 & 73.3 & 46.0 & 59.5
          & 1949 & 5789 & 3080 & 3488 & 3577 \\
      \midrule
      \bottomrule
      \end{NiceTabular}%
    }
    \label{tab:main_results}
    % \vspace{-15pt}
  \end{table}

\subsection{Experimental Setup}
% \jh{no need more multiple paragraphs, just merging them, maybe just two paragraphs, one is setup one is baseline}
\paragraph{Setup} We experiment with three capable and representative LRMs across three different sizes known for their overthinking tendencies: \dss{}, \dsm{} and \dsl{}. We adhere to the original prompt from DeepSeek-R1~\citep{deepseekai2025deepseekr1incentivizingreasoningcapability}, with the full prompt available in Appendix~\ref{ap:training_prompt}. We train these models using the DeepScaleR-Preview-Dataset~\citep{deepscaler2025}, a high-quality mathematics dataset containing 40K competition-level question-answer pairs. We evaluate the models on four benchmarks of varying difficulty: MATH500~\citep{hendrycks2021measuring}, OlympiadBench~\citep{he-etal-2024-olympiadbench}, AIME 2024, and AMC 2023. We set $\alpha=0.5$ for our methods in all experiments to balance the trade-off between correctness rewards and solution length penalties. $L_T$ is a hyper-parameter for our approaches because the automatic adapting mechanism will enumerate the target length from $L_T$ to the context window size to select the adaptive target lengths $L_A$, as described in \S\ref{sec:adaptauto}. 
% that defines the minimum target length, which helps control the trade-off between accuracy and token usage, as lower values of $L_T$ encourage more concise responses. 
Parameter settings for baseline methods are provided in Appendix~\ref{ap:params_rewards}, and full details of our training procedure and evaluation methodology can be found in Appendix~\ref{ap:training_evaL_Aetails}.

\paragraph{Baselines} According to Table~\ref{tab:unified_reward}, we train models using different types of length rewards design and compare our \mone, \mtwo, \mthree{} to previous works. Considering the high computational cost of RL training, we select Efficient Reason~\citep{arora2025traininglanguagemodelsreason} and L1-Max~\citep{aggarwal2025l1controllinglongreasoning} as the representatives, since they perform better accuracy compared to other methods inside same group and are more close to our settings. For ThinkPrune~\citep{hou2025thinkprunepruninglongchainofthought}, we re-evaluate their open-sourced models.

\subsection{Efficacy-Efficiency Trade‑off}
Since there is a trade-off between accuracy and response length, one of the best ways to evaluate different methods is to compare their Pareto-optimal frontiers. We start with the DeepSeek-R1-Distill-Qwen-1.5B model as its small size allows us to run multiple experiments to investigate the trade-off of different approaches. To fully evaluate the potential of each method, we adjust key parameters ($\alpha$ for group-based reward, $L_T$ for other methods) to explore different tradeoffs along the accuracy-length trade-off curve. The full details of different hyper-parameters for different methods can be found in Table~\ref{tab:parms_rewards}. As a result, each point in Figure~\ref{fig:pareto} and Figure~\ref{fig:pareto_full} represents a separate experiment with a fully trained model using a distinct hyperparameter configuration. We also list the results in different benchmarks in Table~\ref{tab:main_results}. Due to the space limit, we leave some results of truncation methods in Table~\ref{tab:main_results_full}.

As shown in Figure~\ref{fig:pareto}, both \mtwo{} and \mthree{} achieve better Pareto-optimal frontiers compared to all other methods. On the AIME2024 benchmark, \mthree{} attains the highest accuracy of $35\%$ using just over $5{,}500$ tokens—a substantial reduction by $63\%$. Meanwhile, \mtwo{} still achieves $34\%$ accuracy with only $4{,}600+$ tokens, underscoring its strong trade-off. Across all benchmarks (Figure~\ref{fig:pareto_full}), \mthree{} achieves the most optimal trade-off when the average token usage is below $3{,}500$, while \mtwo{} performs the best in higher token regimes. Specifically, \mtwo{} achieves $60.3\%$ accuracy with only $3{,}520$ tokens on average, representing a substantial reduction from the $10{,}177$ tokens used by the original model. Compared to the \mone{} method, both \mtwo{} and \mthree{} achieve significant improvements, demonstrating that incorporating a \textbf{dynamic} and \textbf{difficulty-aware} mechanism greatly enhances the efficacy-efficiency trade-off. Compared to other baseline methods, \mone still exhibits a more favorable trade-off.

% The different behaviors of \mtwo{} and \mthree{} suggest that allowing the model to explore incorrect responses by generating longer outputs is beneficial in two scenarios: (1) for harder problems where extended reasoning is often required, and (2) when the target length $L_T$ is set to a small value, where permitting longer incorrect outputs helps mitigate the performance degradation caused by aggressive length decreases. This strategy thus contributes to achieving a better trade-off between accuracy and efficiency.

\begin{table}[t!]
  \centering
  \caption{Accuracy (\%) with average token usage for each dataset and different methods using 7B and 32B models. "Original" denotes the original model. $T_k$ is the truncation method with context window $k$.}
  \resizebox{\textwidth}{!}{%
    \begin{NiceTabular}{lccccc|ccccc}[code-before=\rowcolors{11-13,16-16}{lightblue}{lightblue}]
    % \begin{tabular}{lccccc|ccccc}
      \toprule
      \midrule
        & \multicolumn{5}{c}{\textbf{Accuracy (\%)}}
        & \multicolumn{5}{c}{\textbf{Generation Length (tokens)}} \\
        & \makecell{MATH\\500} & AIME  & AMC   & \makecell{Olympiad\\Bench} & Avg.
        & \makecell{MATH\\500} & AIME  & AMC   & \makecell{Olympiad\\Bench} & Avg. \\
      \midrule
    \multicolumn{11}{c}{\dsm{}} \\
    \midrule
    Original
    & 92.6 & 53.1 & 88.4 & 58.9 & 73.3
    & 4017 & 13414 & 6433 & 8987 & 8213 \\
    $T_{8192}$
    & 92.0 & 51.9 & 88.3 & 56.4 & 72.2
    & 1972 & 5655 & 3159 & 3606 & 3598 \\
    Group
    & 89.4 & 48.1 & 82.8 & 53.7 & 68.5
    & 780 & 4271 & 1693 & 2348 & 2273 \\
    \mone 
    & 92.2 & 54.4 & 89.7 & 58.1 & 73.6
    & 2317 & 6320 & 3733 & 4262 & 4158 \\
    \mtwo
    & 92.2 & 58.3 & 90.0 & 61.0 & 75.4
    & 1836 & 5379 & 2694 & 3350 & 3315 \\
    $\text{\mthree}$
    & 92.0 & 55.8 & 89.1 & 58.9 & 74.0
    & 1658 & 4969 & 2612 & 3157 & 3099 \\
    \midrule
    \multicolumn{11}{c}{\dsl{}} \\
    \midrule
    Original
    & 94.4 & 71.7 & 93.1 & 64.6 & 80.95
    & 3553 & 10335 & 6177 & 7697 & 6941 \\
    \mthree
    & 93.2 & 70.8 & 93.1 & 62.2 & 79.83
    & 2314 & 6785 & 3545 & 4608 & 4313 \\
    \midrule
    \bottomrule
    \end{NiceTabular}%
    % \end{tabular}
  }
  \label{tab:larger_sizes}
  % \vspace{-15pt}
\end{table}

\subsection{Experiments on Larger Models}
\label{sec:larger_sizes}

To better evaluate the effectiveness of our proposed methods,\mone, \mtwo, and \mthree. We conduct experiments on \dsm{}, as shown in Table~\ref{tab:larger_sizes}. Given the computational cost of larger models, we set key hyperparameters for each method to achieve an appropriate trade-off. Specifically, we set $\alpha=0.2$ for the group-based reward, $L_T=8192$ for the truncation method in \mone, $L_T=4096$ for \mtwo{} and \mthree. Notably, we do not tune $\alpha$ with fixed value $0.5$ in all experiments of our methods. As shown in Table~\ref{tab:larger_sizes}, \mtwo{} achieves the best trade-off with better accuracy and significantly fewer tokens. On the AIME dataset, it achieves an accuracy of $58.3\%$, representing a gain of $+\mathbf{5.2}$ points, while using only $5,379$ tokens—substantially fewer than the $13,414$ tokens used by the original model. Compared to other methods, \mone{}, \mtwo{}, and \mthree{} also attain better trade-offs on most benchmarks, particularly on the more challenging ones.

% To better evaluate the effectiveness of our method, we scale the model size to 32B, as shown in Table~\ref{tab:larger_sizes}. 
For the 32B model, due to computational constraints, we compare the \mthree-trained model with the original baseline under this larger setting and set $L_T = 8192$. \mthree{} achieves competitive accuracy with only a minor drop ($1\%$), while still significantly reducing output length. Notably, the accuracy of \dsl{} on our training dataset is already very high—over $76\%$, leaving little room for further improvement. We speculate that with more challenging and diverse training data, \mthree{} could yield further accuracy gains.

\subsection{Experiments on Out-of-Domain Benchmarks}
\label{sec:ood}

\begin{figure}[t!]
  \centering
  \begin{subfigure}[t]{0.48\textwidth}
      \centering
      \includegraphics[width=\textwidth]{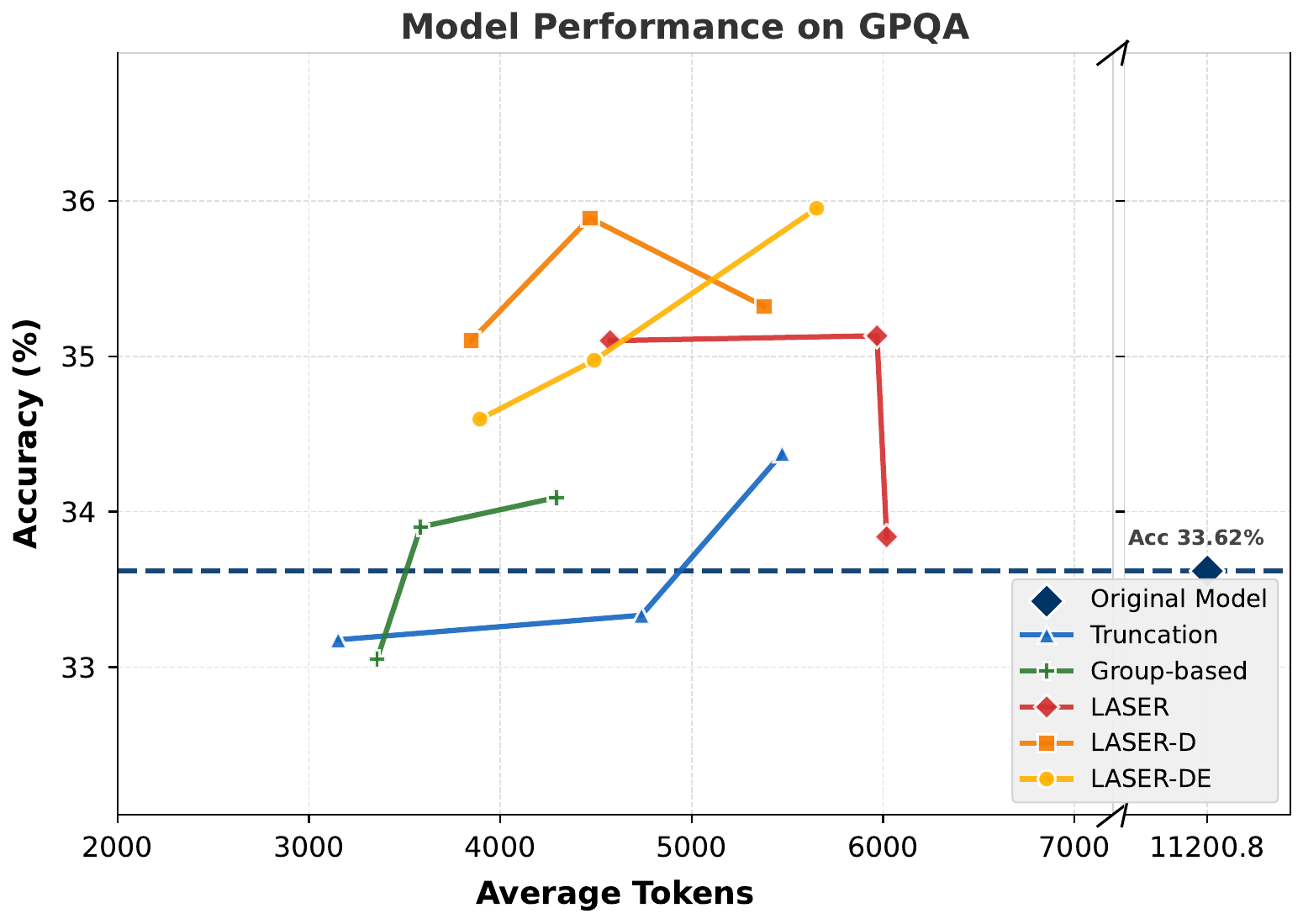}
      \label{fig:ood-gpqa}
  \end{subfigure}
  \hfill
  \begin{subfigure}[t]{0.48\textwidth}
      \centering
      \includegraphics[width=\textwidth]{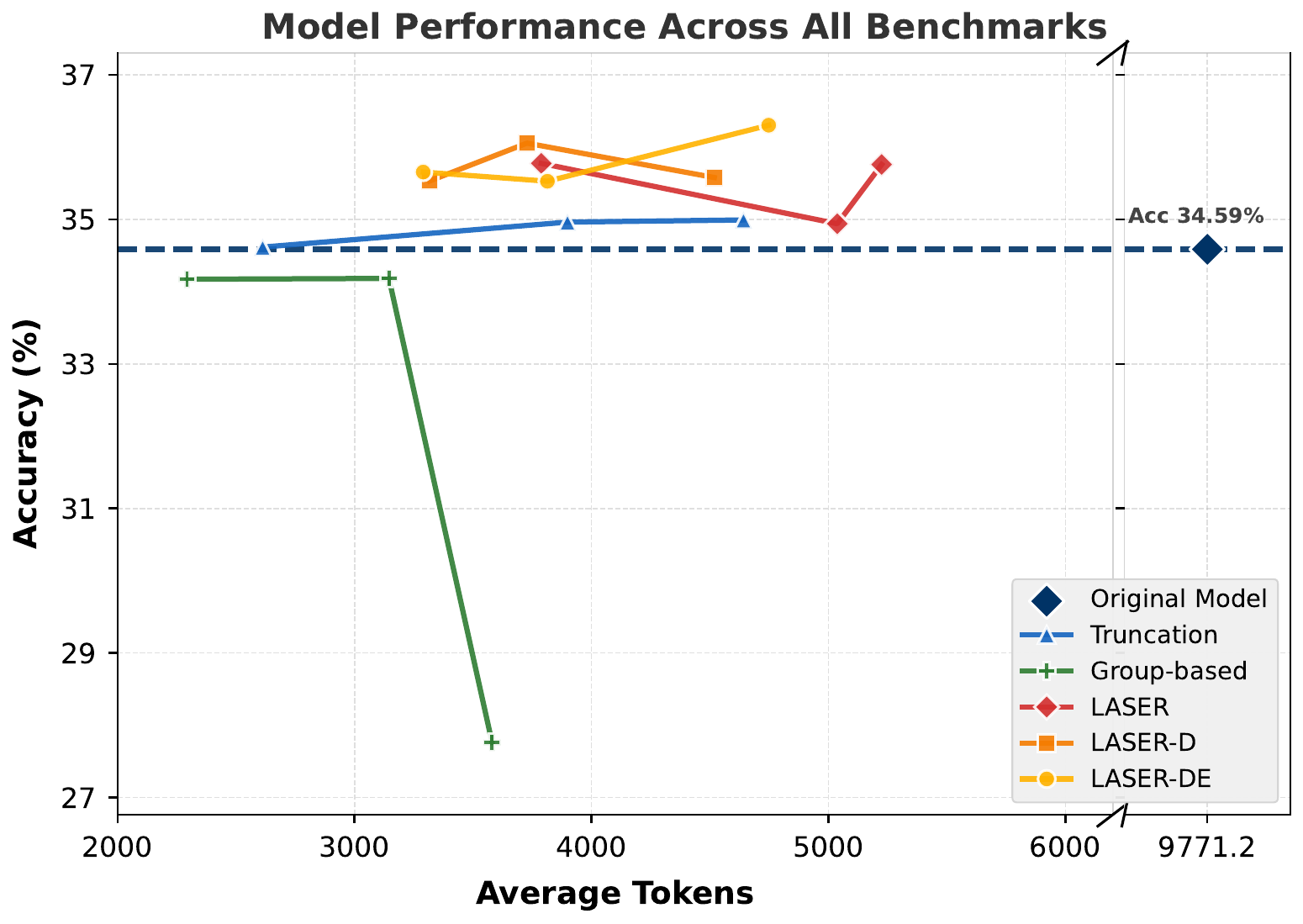}
      \label{fig:ood-average}
  \end{subfigure}
  \caption{Performance on out-of-domain benchmarks: GPQA and average performance across all three benchmarks (GPQA, MMLU, LSAT).}
  \label{fig:ood-performance}
\end{figure}

We evaluate whether \mone, \mtwo{} and \mthree{} can generalize to domains outside the RL training distribution. We select three out-of-domain benchmarks: GPQA~\citep{rein2023gpqagraduatelevelgoogleproofqa}, LSAT~\citep{wang2022lsat}, and MMLU~\citep{hendrycks2021measuringmassivemultitasklanguage}, following the evaluation settings established by L1~\citep{aggarwal2025l1controllinglongreasoning}. Figure~\ref{fig:ood-performance} illustrates the efficacy-efficiency trade-off on GPQA and the average performance across all benchmarks. Compared to the original model, \mone, \mtwo{} and \mthree{} consistently achieve significant improvements in both accuracy and token usage, demonstrating robust generalization capabilities. And \mtwo{} and \mthree{} maintain the best trade-off even when compared to \mone.
% When considering the average across all benchmarks, however, the trade-offs between \mone{}, \mtwo{} and \mthree{} become more competitive. We speculate that this convergence is due to knowledge-focused questions like those in MMLU, which appear to benefit less from extended reasoning traces.

\section{Analysis}
\label{sec:analysis}
We use \dss{} as the backbone, conduct a comprehensive analysis that includes budget-forcing inference, dynamics of adaptive target lengths, shifts in reasoning patterns, and qualitative evaluations. Due to the space limit, please refer to Appendix~\ref{ap:budget_forcing} for budget-forcing inference, Appendix~\ref{ap:dynamics_target_length} for dynamics of adaptive target lengths.

\subsection{Changes of Thinking Patterns} To better understand the changes of response length, we analyze the changes of thinking patterns over RL iterations on AIME2024 with 16 samples per question. We analyze through two approaches, keywords counts~\citep{yeo2025demystifyinglongchainofthoughtreasoning} and reasoning behavior ratios~\citep{zeng2025simplerlzooinvestigatingtamingzero}.

\begin{figure}[t!]
  \centering
  \includegraphics[width=\textwidth]{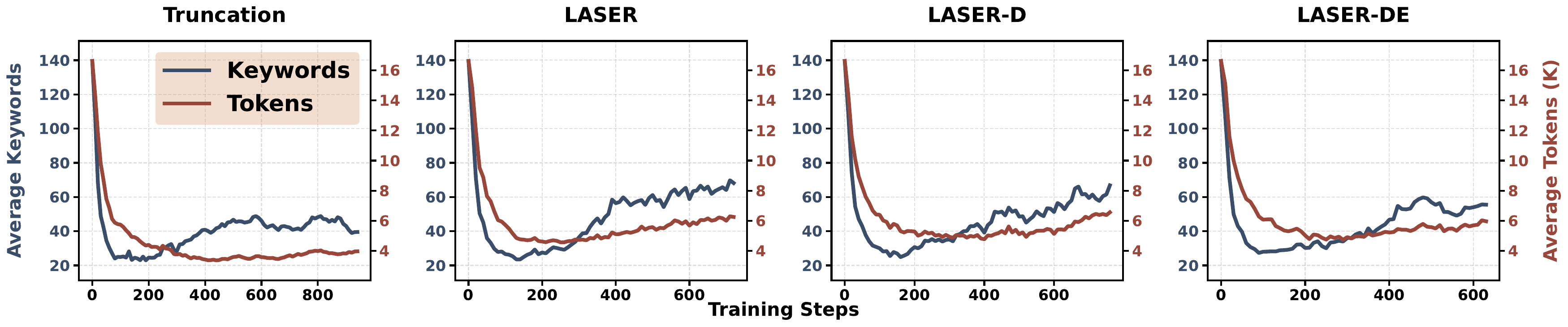}
  \vspace{-5pt}
  \caption{Average keyword amount and response length over RL training on AIME24. The truncation method uses a 8192 token context window. \mone, \mtwo, and \mthree{} employ a target length of $L_T=2048$.}
  \label{fig:keywords} \vspace{-13pt}
\end{figure}

\paragraph{Shifts in ``Self-Reflection'' Keywords}  
``Self-reflection'' or ``Aha-moment'' reasoning has emerged as an intriguing behavior in LRMs~\citep{deepseekai2025deepseekr1incentivizingreasoningcapability}. Following previous work~\citep{yeo2025demystifyinglongchainofthoughtreasoning}, we track this behavior by monitoring seven representative keywords: [\emph{``recheck''},\emph{``rethink''},\emph{``try again''},\emph{``wait''}, \emph{``alternatively''},\emph{``retry''},\emph{``however''}]. As shown in Figure~\ref{fig:keywords}, the average amount of these keywords (occurrences per token) declines notably as response length decreases across all methods. This suggests that RL may reduce instances of spurious ``self-reflection,'' previously identified as a contributor to over-thinking~\citep{chen2025think23overthinkingo1like}. Interestingly, as training progresses, we observe increased keyword amount while maintaining shorter outputs, indicating models develop more efficient self-reflection behaviors without producing verbose responses.

\paragraph{Changes in Thinking Behaviors}  
To further investigate the changes of reasoning patterns beyond keyword statistics, we employ \texttt{gpt-4.1-mini} to perform a more fine-grained analysis of cognitive behaviors throughout the training process. Specifically, we adopt the cognitive behavior framework proposed by~\citep{gandhi2025cognitivebehaviorsenableselfimproving}, which identifies reasoning-related behaviors such as \emph{Backtracking}, \emph{Verification}, \emph{Subgoal Setting}, and \emph{Enumeration}. We report the proportion of each behavior relative to the total number of behaviors, focusing on these four representative categories. The complete list of behaviors and implementation details are provided in Appendix~\ref{ap:behaviors}.

As shown in Figure~\ref{fig:behaviors}, the proportion of \emph{Backtracking} behavior decreases significantly, from over $30\%$ to just above $10\%$, as the response length is reduced. This trend aligns with the keyword statistics, as many of the tracked keywords (e.g., ``recheck'', ``retry'', ``rethink'') are indicative of \emph{Backtracking}. While \emph{Backtracking} declines during training, the proportions of other reasoning behaviors, \emph{Verification}, \emph{Enumeration}, and \emph{Subgoal Setting}, remain stable, with a slight increase observed in \emph{Subgoal Setting}. These results suggest that reducing response length does not degrade the model into a non-reasoning baseline. On the contrary, core reasoning behaviors are preserved, while unnecessary backtracking is minimized, indicating more efficient reasoning in the refined models.

\subsection{Qualitative Analysis}
We conduct a qualitative analysis on the trivial question ``1+1=?'' and the MATH500 dataset to understand how RL improves reasoning efficiency. Comparing the original \dss{} model with the \mtwo-trained version, Figure~\ref{fig:pareto} illustrates how the original model generates repetitive ``self-reflection'' even for trivial questions, while the trained model directly provides the answer. Our analysis of MATH500 (detailed in Appendix~\ref{ap:case_study}) reveals that the original model tends towards verbose, redundant explanations of single ideas. In contrast, the \mtwo-trained model expresses the same concepts more succinctly using structured formulas, significantly improving token efficiency. This suggests our RL-based approach not only reduces unproductive backtracking but also encourages a shift towards more concise and direct expression.

\begin{figure}[t!]
  \centering
  \includegraphics[width=\textwidth]{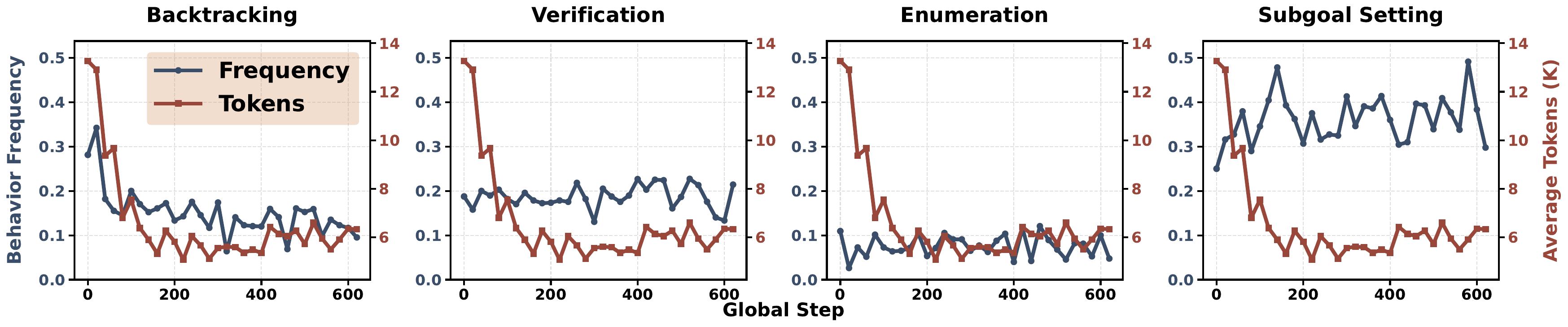}
  \vspace{-5pt}
  \caption{Changes in reasoning behaviors ratio and response length over RL training iterations on AIME2024. The figure shows how \mthree's thinking patterns change during training with a target length of $L_T=2048$.} \vspace{-15pt}
  \label{fig:behaviors}
\end{figure}

\section{Conclusion}
In this paper, we 
% investigate RL-based CoT compression, starting by a simple yet effective baseline, truncation. To better understand it, we 
propose a unified view for RL-based CoT compression, unifying various reward-shaping and truncation methods. Building on this view, we introduce new approaches with adaptive, length-based reward shaping.
% \mone, a novel reward-shaping method that employs a step reward function guided by a target length. To further optimize the efficacy-efficiency trade-off, we present a dynamic and difficulty-aware length-based step reward (\mtwo) method and its variant, \mthree, which dynamically train data of varying difficulty levels with different target lengths. 
Extensive experiments demonstrate that our proposed methods achieve superior Pareto-optimality and significant improvements in both accuracy and token efficiency. Our analysis of reasoning behaviors reveals that our RL-based CoT compression effectively encourages models to reason more concisely and productively.

\bibliographystyle{abbrvnat}  %plainnat,abbrvnat,unsrtnat
\small
\bibliography{Reference}
\normalsize

\medskip

%%%%%%%%%%%%%%%%%%%%%%%%%%%%%%%%%%%%%%%%%%%%%%%%%%%%%%%%%%%%

\newpage

\appendix
\section{Pareto-Optimality}
\label{ap:full_pareto_figures}
\begin{figure}[htbp]
    \centering
    \begin{subfigure}[t]{0.48\textwidth}
        \centering
        \includegraphics[width=\textwidth]{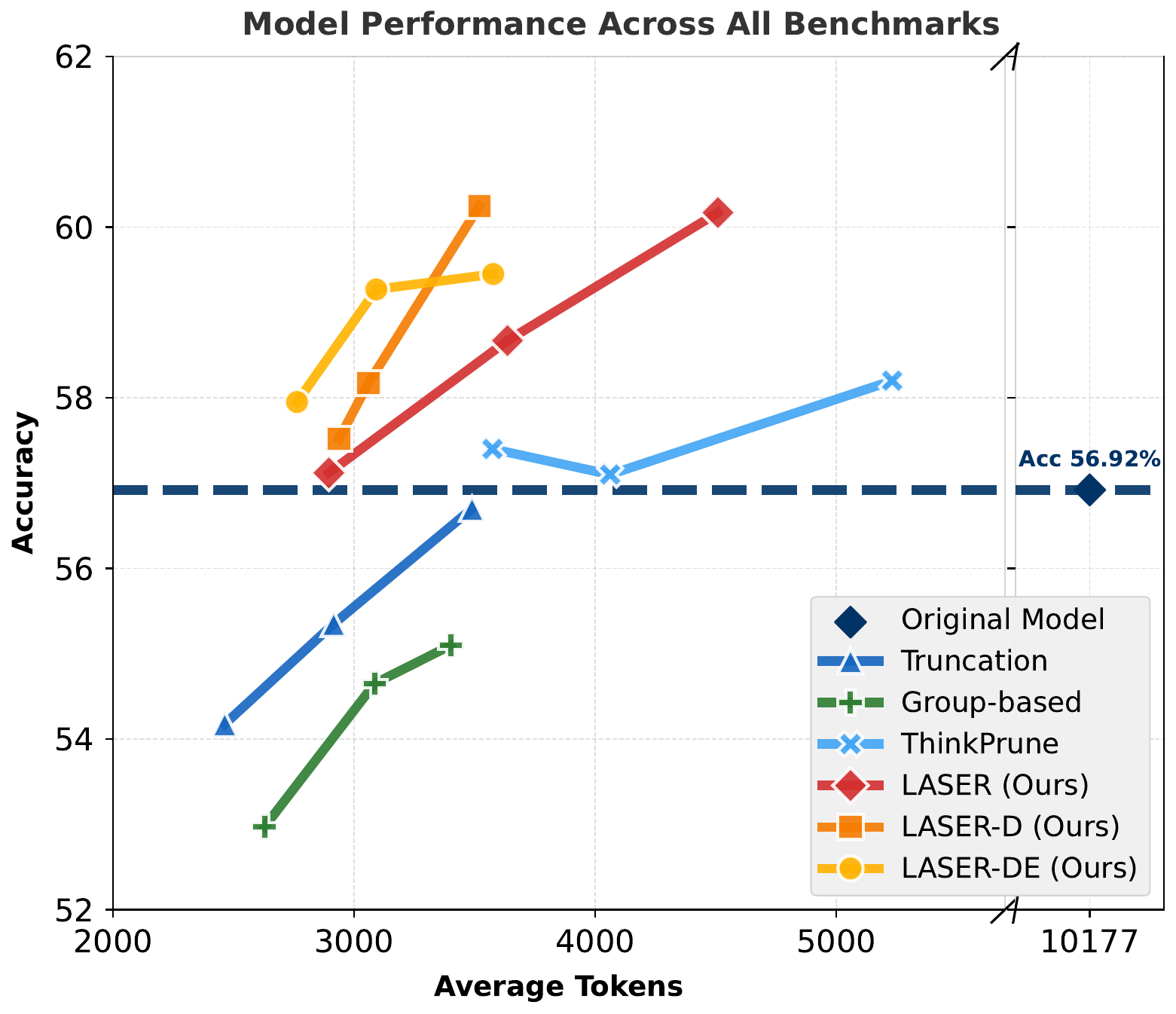}
        \caption{}
    \end{subfigure}
    \hfill
    \begin{subfigure}[t]{0.48\textwidth}
        \centering
        \includegraphics[width=\textwidth]{figures/average_aime.pdf}
        \caption{}
    \end{subfigure}
    \caption{Pareto-optimal trade-off between accuracy and response length across various methods. Each point represents a single training run with different hyper-parameters. Our methods, \mthree, \mtwo, and \mone, achieve a Pareto-optimal trade-off compared to all other methods. (a) Accuracy and response length on all benchmarks (MATH500, AIME2024, AMC2023, Olympiad Bench) (b) Accuracy and response length on AIME2024}
    \label{fig:pareto_full}
\end{figure}

We illustrate the efficacy-efficiency trade-off in Figure~\ref{fig:pareto_full}. Our proposed methods, \mone, \mtwo, and \mthree, demonstrate significant improvements in both accuracy and token usage across all benchmarks, particularly in the most challenging ones. Notably, \mtwo{} and \mthree{} achieve a Pareto-optimal trade-off compared to all other methods.

\section{Ratio of Truncated Responses During Training with Truncation}

We analyze the ratio of truncated responses when applying an 8192 token limit during training. Our findings show that the proportion of truncated responses is initially very high—exceeding 45\%, and remains substantial (above 10\%) even after 200 rollout steps. This high truncation rate highlights the context window constraints in training is sub-optimal.

\begin{figure}[htbp]
  \centering
  \includegraphics[width=0.6\textwidth]{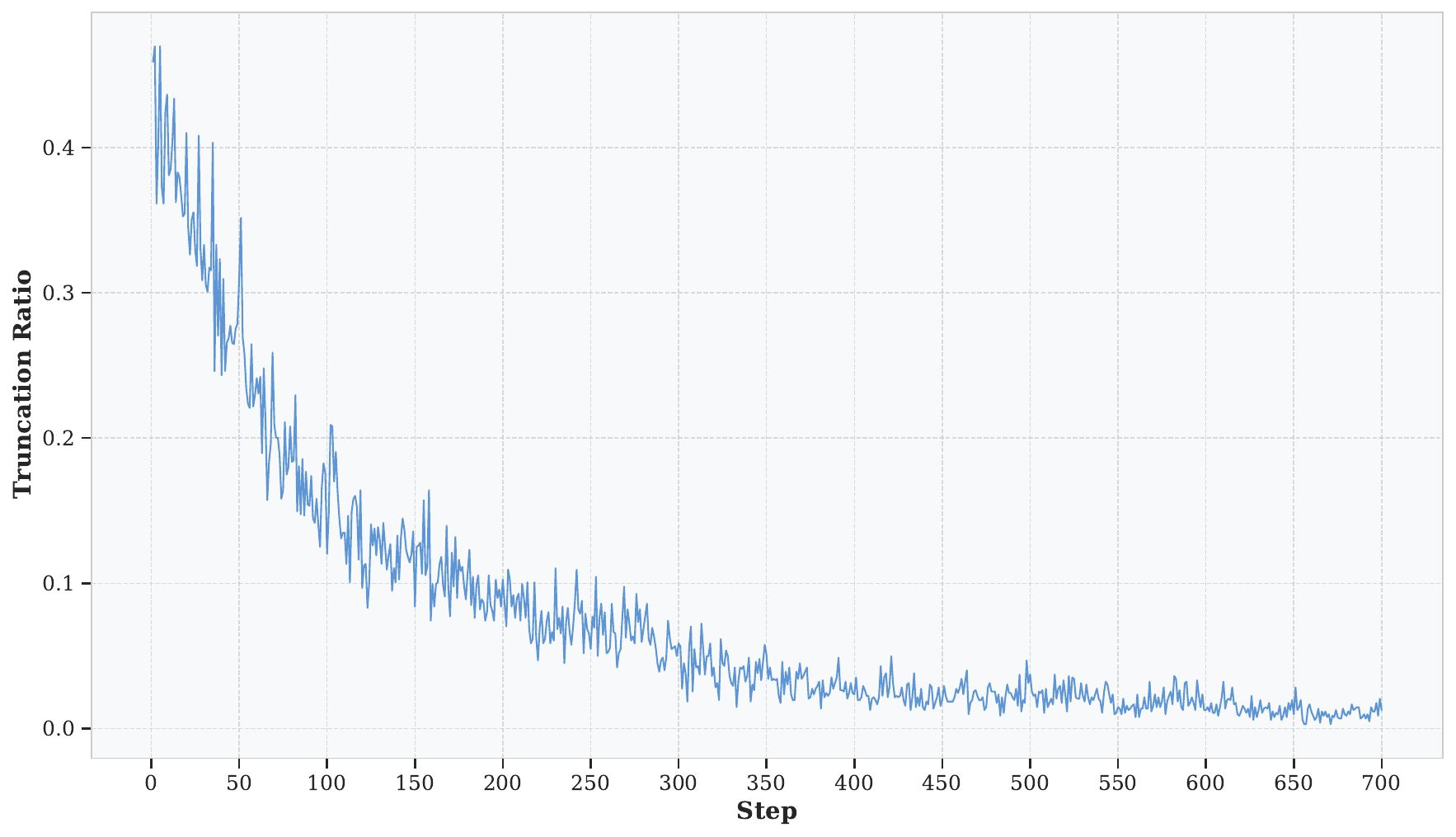}
  \caption{The ratio of truncated responses in training data with 8192 tokens limit.}
  \label{fig:clipratio1}
\end{figure}

\section{Dynamics of Accuracy and Rewards Throughout Training}
\label{ap:acc_rewards}
\begin{figure}[t!]
  \centering
  \begin{subfigure}[t]{0.48\textwidth}
      \centering
      \includegraphics[width=\textwidth]{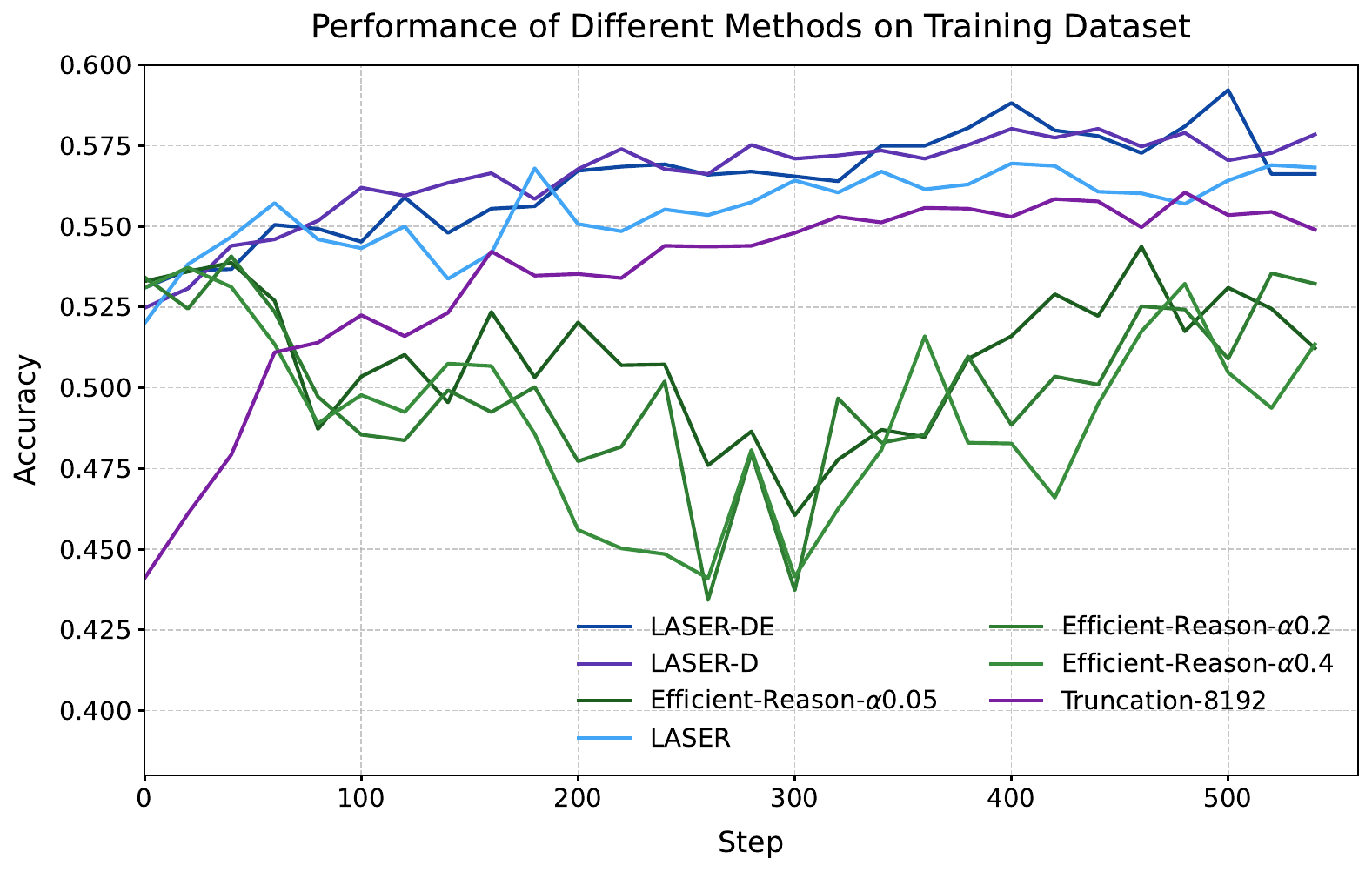}
      % \caption{Accuracy on training dataset across training iterations for different methods}
      \caption{}
      \label{fig:training_accuracy}
  \end{subfigure}
  \hfill
  \begin{subfigure}[t]{0.48\textwidth}
      \centering
      \includegraphics[width=\textwidth]{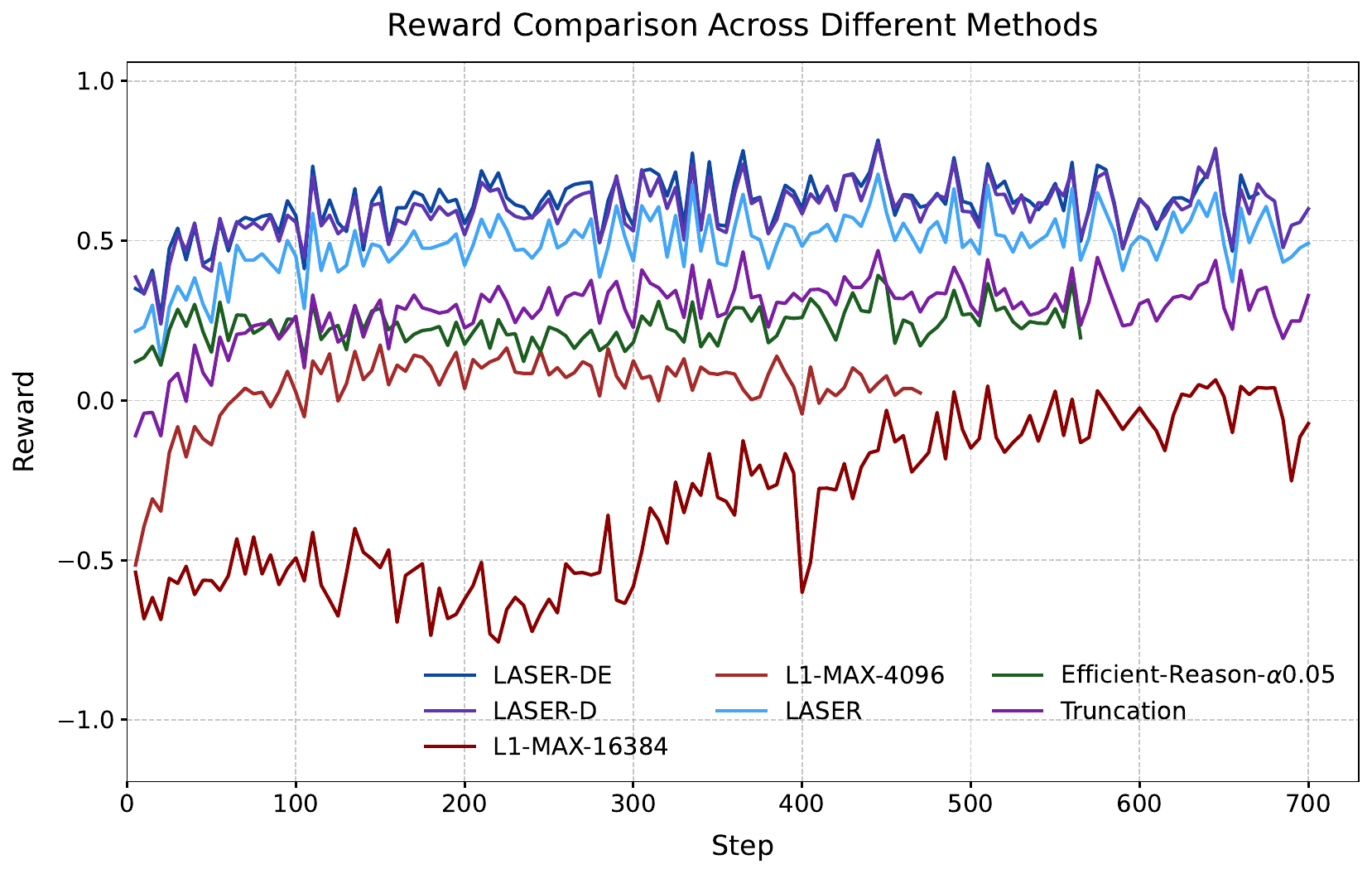}
      % \caption{Rewards across training iterations for different methods}
      \caption{}
      \label{fig:reward_comparison}
  \end{subfigure}
  % \caption{Accuracy and rewards on training dataset during training process.}
  \caption{(a) Accuracy on training dataset across training iterations for different methods (b) Rewards across training iterations for different methods}
  \label{fig:combined_results}
\end{figure}

We present the accuracy and rewards for various methods across training iterations in Figure~\ref{fig:training_accuracy} and Figure~\ref{fig:reward_comparison}. As discussed in \S\ref{sec:unified}, group-based rewards tend to exploit the length rewards $S(y)$ while causing a significant drop in accuracy. Budget-based rewards such as L1-Max-16384~\citep{aggarwal2025l1controllinglongreasoning} suffer from unstable training when the context window is large. In contrast,  other methods like truncation methods, \mone, \mtwo, and \mthree{} demonstrate a simultaneous increase in both rewards and accuracy throughout the training process.

\section{Supplementary Details: Length-based Reward Shaping Formulations}
\label{ap:details_unified}
In this section, we provide additional details regarding the various formulations of length-based reward shaping as presented in Table~\ref{tab:unified_reward}. These formulations can viewed as different variants of our unified framework in Eq.~\ref{eq:unified_reward} which can be implemented by making specific design choices for three key components: $\csymbol{}$, $\lambda(y)$, and $S(y)$ inside the framework. Here we review the formulation of Eq.~\ref{eq:unified_reward} to better illustrate following approaches.

\begin{equation*}
  \hat{R}(x, y) = C(y) + \lambda(y)\cdot S(y)
\end{equation*}

\subsection{Truncation}
\paragraph{Vanilla Truncation}
As aforementioned discussions (\S\ref{sec:unified}), truncation is a special case of the length reward with \(\csymbol{}=0\), where the target length \(L_T\) is enforced by the context window. $\rho$ is set as $0$. It follows the design:

\begin{align*}
    C(y) &= 0 \\
    \lambda(y) &= 1 \\
    S(y) &= \begin{cases}
            R(x,y) & \text{if } L(y) \leq L_T \\
            \rho & \text{if } L(y) > L_T
          \end{cases}
\end{align*}

\paragraph{ThinkPrune}
ThinkPrune~\citep{hou2025thinkprunepruninglongchainofthought} is another truncation-based approach, which extends vanilla truncation by introducing an adaptive target lengths \(L_A\) to replace fixed target lengths \(L_T\). $\rho$ is set as $0$. The design follows:

\begin{align*}
    C(y) &= 0 \\
    \lambda(y) &= 1 \\
    S(y) &= \begin{cases}
      R(x,y) & \text{if } L(y) \leq L_A \\
      \rho & \text{if } L(y) > L_A
    \end{cases}
\end{align*}

Their training methodology employs a progressive three-stage process with iterative refinement of $L_A$. Each subsequent stage initializes from the checkpoint of the previous stage while mannually reducing the value of $L_A$. Specifically, they progressively decrease \(L_A\) through values of $4096$, $3072$, and $2048$ across the three stages.

\subsection{Group-based Rewards}
In the context of group-based rewards, the length reward $S(y)$ is specifically designed to promote brevity by assigning higher scores to shorter responses within a rollout group. This mechanism functions as a comparison-based reward system that inherently favors more concise responses. Most of them follow the design $C(y)=R(x,y)$ to keep the accuracy performance of models.

\paragraph{Efficient Reasoning}
Efficient Reasoning~\citep{arora2025traininglanguagemodelsreason} follows the principle of group-based reward by specifically encouraging conciseness within correct responses. The mean and variance scalars are computed exclusively from the subset of correct responses, ensuring appropriate statistical distributions. By selectively rewarding conciseness only when answers are correct, this approach maintains higher accuracy compared to Kimi-k1.5~\citep{kimiteam2025kimik15scalingreinforcement}, which encourages wrong responses to be shorter. Considering the similarity between the two approaches and the better efficacy-efficiency trade-off, we select Efficient Reasoning as the representative group-based reward in this paper. The corresponding design can be formulated as follows:

\begin{align*}
    C(y) &= R(x,y) \\
    \lambda(y) &= \mathbb{I}(R) \\
    S(y) &= -\alpha \cdot \sigma\left(\frac{L(y) - Mean(y)}{STD(L)}\right)
\end{align*}

\paragraph{Kimi-k1.5}
The design of Kimi-k1.5 is similar to Efficient Reasoning~\citep{arora2025traininglanguagemodelsreason}, with two main differences. First, the scalar factors are computed using the minimum response length and the difference between maximum response length and maximum length within a rollout group. Second, Kimi-k1.5 encourages all responses to be shorter, rather than focusing solely on shortening correct responses. Such a design has the potential to intensify reward hacking, as models may exploit the reward function by favoring shorter responses to maximize their scores. The designs follows:

\begin{align*}
    C(y) &= R(x,y) \\
    \lambda(y) &= 1 \\
    S(y) &= \begin{cases}
      0.5-\tfrac{L(y)-L_{\min}}{L_{\max}-L_{\min}} & \text{if } \mathbb{I}(R) = 1 \\
      \min\!\left(0,\;0.5-\tfrac{L(y)-L_{\min}}{L_{\max}-L_{\min}}\right) & \text{if } \mathbb{I}(R) = 0
    \end{cases}
\end{align*}

\subsection{Budget-based Reward}
Budget-based rewards use query-specific target lengths (budgets) and penalize responses that deviate from these instructions. And the coefficient \(\alpha\) controls the trade-off between length reward and correctness reward. They come in two flavors: exact mode and max mode. We follow same settings as L1~\citep{aggarwal2025l1controllinglongreasoning} and set $\alpha=0.0003$ for exact mode, $\alpha=0.01$ for max mode. 

\paragraph{Exact Mode}
In exact mode, the model must hit the specified target length $L_T$ exactly, and any deviation (even shorter outputs) is penalized. The design can be formulated as:

\begin{align*}
  C(y) &= R(x,y) \\
  \lambda(y) &= 1 \\
  S(y) &= -\alpha\cdot |L(y) - L_T|
\end{align*}

\paragraph{Max Mode}
In max mode, only outputs that exceed $L_T$ incur a penalty. The designs follow:

\begin{align*}
  C(y) &= 0 \\
  \lambda(y) &= \mathbb{I}(R) \\
  S(y) &= \operatorname{clip}(\alpha \cdot (L(y) - L_T) + \delta, 0, 1)
\end{align*}

\section{Training Configurations}
We leverage the prompt from~\citet{deepseekai2025deepseekr1incentivizingreasoningcapability}, which is shown in Figure~\ref{fig:training_prompt}. And the mark for thinking is ``<think>...</think>''.

\subsection{Training Prompt}
\label{ap:training_prompt}
\begin{figure}[htbp]
  \centering
  \includegraphics[width=0.8\textwidth]{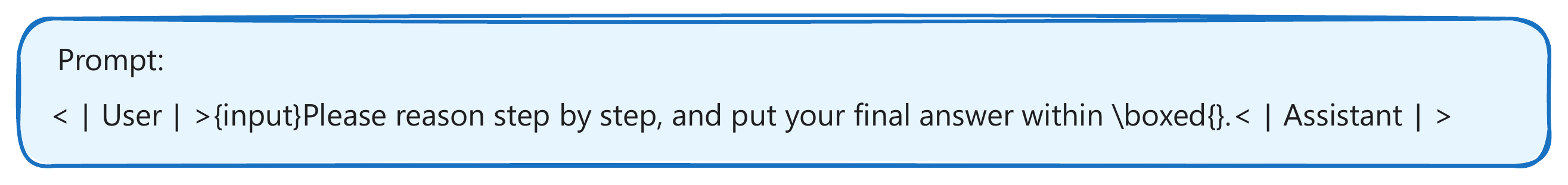}
  \caption{Training prompt for our training.}
  \label{fig:training_prompt}
\end{figure}

We list our training prompt in Figure~\ref{fig:training_prompt}, which follows the prompt from DeepSeek-R1~\citep{deepseekai2025deepseekr1incentivizingreasoningcapability}.

\subsection{Training and Evaluation Details}
\label{ap:training_evaL_Aetails}
We employ the verl~\citep{sheng2024hybridflow} framework for model training and Qwen-Math-Eval~\citep{yang2024qwen25mathtechnicalreportmathematical} for evaluation. During training, we set the rollout batch size to 128, conduct 8 rollouts per prompt, use a temperature of 0.6, and train with a mini-batch size of 64. In our preliminary experiments, we found long-to-short RL benefits from clip-higher strategy~\citep{yu2025dapoopensourcellmreinforcement}. So we follow DAPO~\citep{yu2025dapoopensourcellmreinforcement} and set $\epsilon_{high}$ as 0.28. For evaluation, we maintain a sampling temperature of 0.6 and permit a maximum of 32,768 tokens to be generated. The number of samplings during evaluation is contingent on the dataset size: 4 samples per question for MATH500 and OlympiadBench, and 16 samples for AIME 2024 and AMC 2023.

\subsection{Full Hyper-Parameter List for Different Length-based Rewards}
\label{ap:params_rewards}
We list the all hyper-parameters for $L_T$, $\alpha$ and $L_A$ in Table~\ref{tab:parms_rewards}.

\begin{table}[htbp]
  \centering
  \caption{The details of key hyper-parameters for different methods}
  \begin{tabular}{c@{\hspace{2cm}}c}
  \toprule
      Methods & Hyper-Parameters \\
      \midrule
      Truncation & $L_T=[10240, 8192, 7168, 6144, 4098, 2048]$ \\
      Think-Prune & $L_A=[4096, 3072, 2048]$ \\
      Group-Based Rewards & $\alpha=[0.4, 0.2, 0.1, 0.05]$ \\
      L1-Max & $\alpha=0.01$ \\
      \mone & $L_T=[8192, 4096, 2048]$ \\
      \mtwo & $L_T=[4096, 2048, 1024]$ \\
      \mthree & $L_T=[4096, 2048, 1024]$ \\
      \bottomrule
  \end{tabular}
  \label{tab:parms_rewards}
\end{table}

\section{Budget-Forcing Inference}
\label{ap:budget_forcing}

\begin{figure}[htbp]
  \centering
  \begin{subfigure}[b]{0.48\textwidth}
      \centering
      \includegraphics[width=\textwidth]{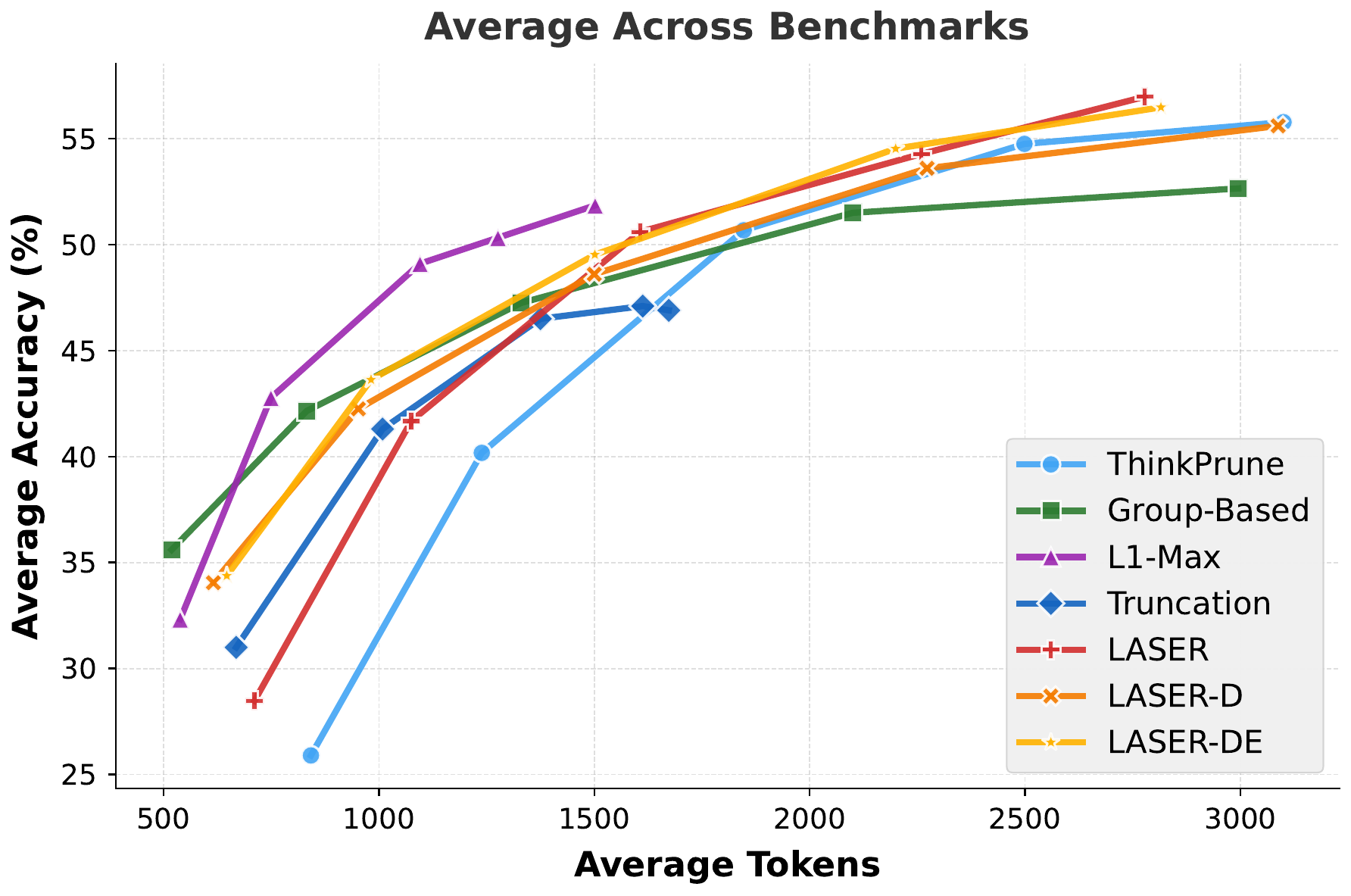}
      \caption{}
      \label{fig:budget_avg}
  \end{subfigure}
  \hfill
  \begin{subfigure}[b]{0.48\textwidth}
      \centering
      \includegraphics[width=\textwidth]{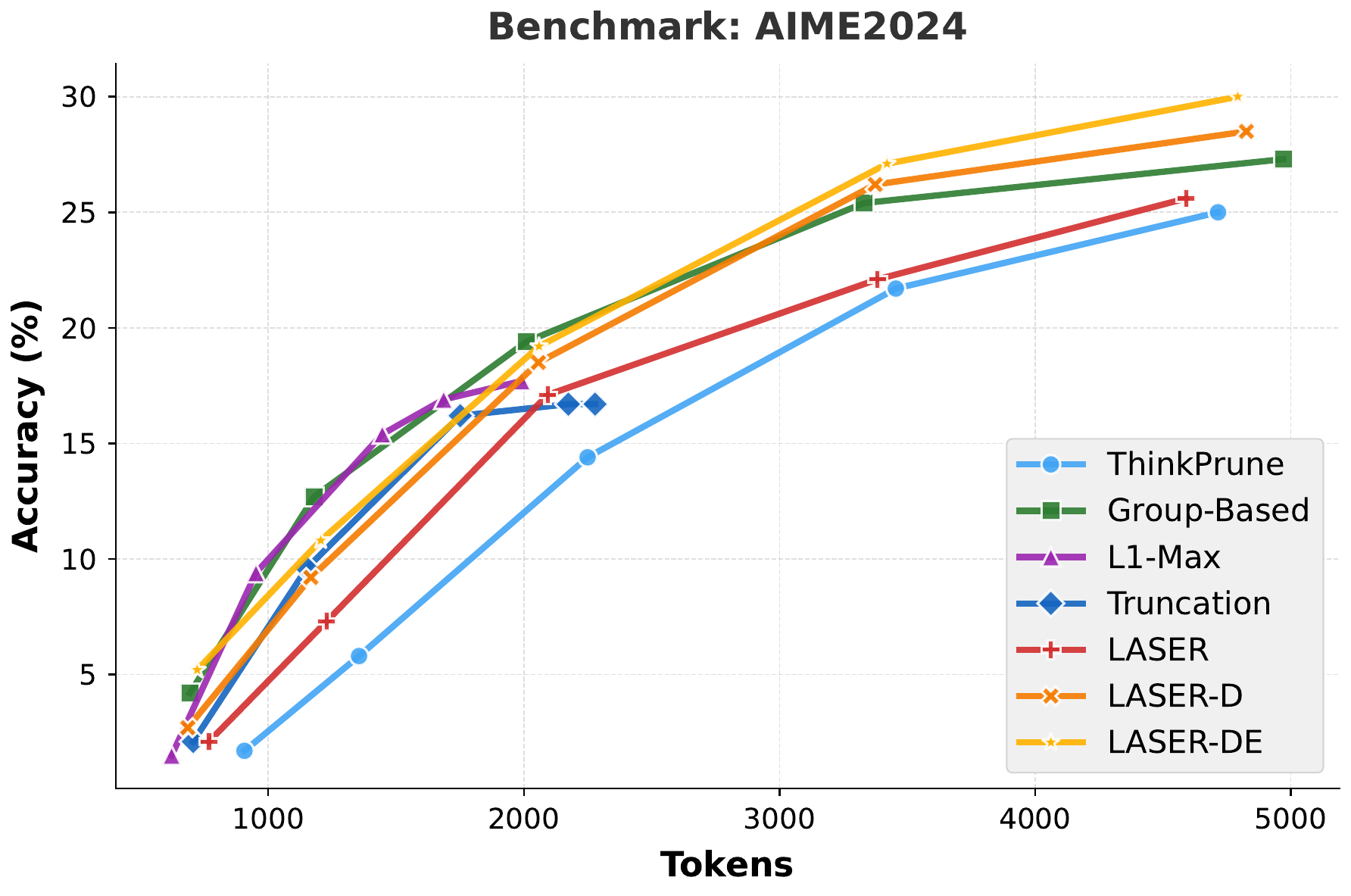}
      \caption{}
      \label{fig:budget_aime}
  \end{subfigure}
  \caption{Budget-forcing inference with different methods. (a) Average accuracy with different output budget on all benchmarks (b) The accuracy of different methods on AIME2024 with different output budget.}
  \label{fig:budget_forcing_analysis}
\end{figure}

To further analyze the impact of different length rewards, we conduct experiments using the budget-forcing setup introduced in S1~\citep{muennighoff2025s1simpletesttimescaling}, which restricts the model to stop reasoning after a fixed number of tokens $B$. We adopt their experimental setting and evaluate across $B = [500, 1000, 2000, 4000, 8000]$ We follow the budget-forcing implementations of \citet{muennighoff2025s1simpletesttimescaling,hou2025thinkprunepruninglongchainofthought}. Specifically, we follow their implementations and modify the codebase of Qwen-Math-Eval. We stop the thinking process of LRMs by appending ``</think>\verb|\n\n|**Final Answer.**''. Since empirically, \dss{} typically summarize its final answer starting with ``\verb|\n\n|**Final Answer.**''. We use the same settings as our evaluations where we sample responses for multiple times with $\operatorname{temperature}=0.6$.. As shown in Figures~\ref{fig:budget_avg} and~\ref{fig:budget_aime}, despite not being explicitly trained with any budget-forcing mechanisms, \mtwo{} and \mthree{} consistently achieve strong trade-offs between accuracy and token efficiency, particularly on harder questions or when inference budgets are moderately constrained.

While \mone{} performs competitively on average benchmarks, it lags behind \mtwo/\mthree{} under strict token budgets or on more challenging examples. L1-Max, specifically trained to meet varying budget constraints during training, performs best under extremely tight budgets, demonstrating the strength of budget-specific optimization. However, its performance plateaus when more budget is available, limiting its ability to improve on harder tasks and resulting in a suboptimal trade-off, as shown in Figure~\ref{fig:budget_aime}. Group-based methods are also effective in low-budget scenarios due to their reward structure favoring shorter outputs, though this often leads to overly brief responses. ThinkPrune shows comparable performance to \mone{} under looser budgets but inherits the limitations of truncation-based approaches, struggling on difficult problems even when more tokens are available.

\section{Dynamics of Adaptive Target Lengths}
\label{ap:dynamics_target_length}
\begin{figure}[htbp]
  \centering
  \includegraphics[width=\textwidth]{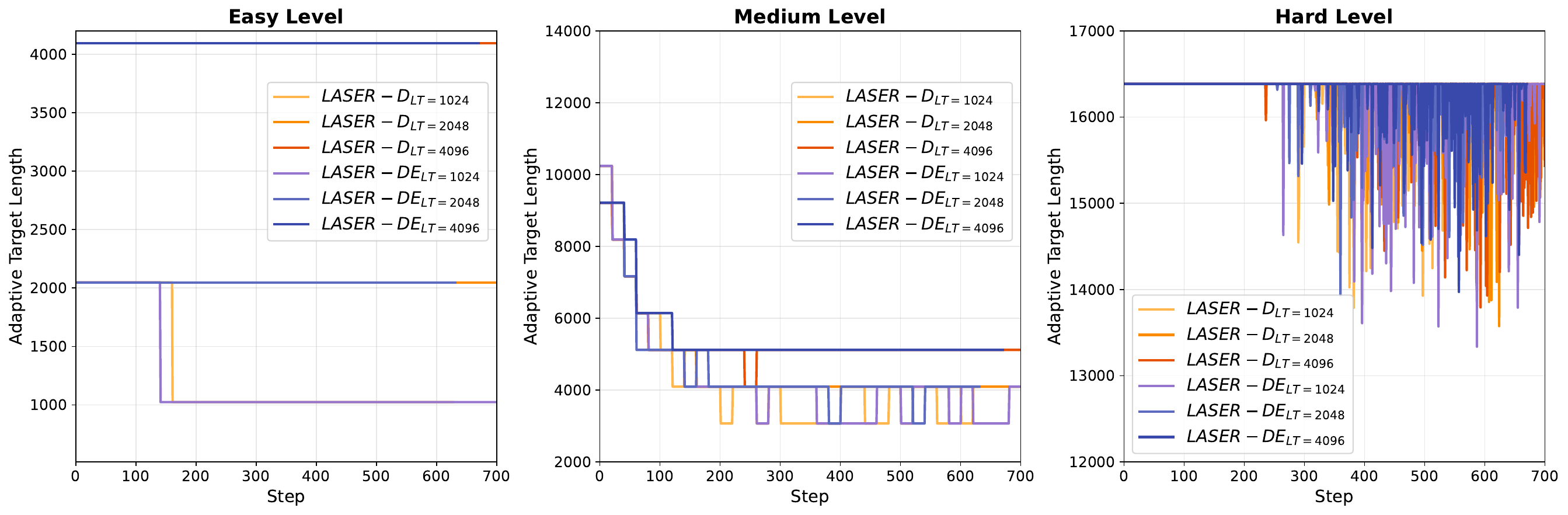}
  \caption{Dynamics of adaptive target lengths during the training process of \mtwo{} and \mthree{}. The figure shows how the adaptive target length $L_A$ changes over training iterations for problems of different difficulty levels (easy, medium, hard). For easy problems, the model selects short target lengths; for medium problems, it gradually decreases from higher initial values; and for hard problems, it maintains consistently high target lengths near the context window limit. This demonstrates the methods' ability to adaptively allocate computational resources based on problem complexity, unlike fixed-length approaches.}

  \label{fig:target_length_dynamics}
\end{figure}

In this section, we analyze the dynamics of adaptive target lengths during the training process of \mtwo{} and \mthree{}. Figure~\ref{fig:target_length_dynamics} shows how the adaptive target length $L_A$ changes over training iterations for both methods.

As demonstrated in Figure~\ref{fig:target_length_dynamics}, our method dynamically selects appropriate target lengths based on problem difficulty. For easy problems (left figure), the model quickly identifies that shorter target lengths are sufficient. For medium-difficulty problems (middle figure), the model begins with longer target lengths (10,000+) and gradually reduces them to 3000-4000 as training continues. For difficult problems (right figure), the model consistently maintains target lengths near the maximum context window size, with some fluctuations attributable to computational precision issues. This adaptive behavior highlights the effectiveness of our approach in efficiently allocating computational resources based on problem complexity.

\section{Full Main Results}
\label{ap:full_results}

We list the full results of different methods in Table~\ref{tab:main_results_full}.

\begin{table}[t!]
    \centering
    \caption{Full results of accuracy (\%) with average token usage for each dataset and different methods. The base model is \dss{}. "Original" denotes the original model. $T_k$ is the truncation method with context window $k$. ``Group'' denotes the Efficient Reasoning~\citep{arora2025traininglanguagemodelsreason} with different $\alpha$. Due to the space limit, we only show three most representative results of truncation method here.}
    \resizebox{\textwidth}{!}{%
    \begin{NiceTabular}{lccccc|ccccc}[code-before=\rowcolors{20-29}{lightblue}{lightblue}]
        \toprule
        \midrule
          & \multicolumn{5}{c}{\textbf{Accuracy (\%)}}
          & \multicolumn{5}{c}{\textbf{Generation Length (tokens)}} \\
          & \makecell{MATH\\500} & AIME  & AMC   & \makecell{Olympiad\\Bench} & Avg.
          & \makecell{MATH\\500} & AIME  & AMC   & \makecell{Olympiad\\Bench} & Avg. \\
        \midrule
        % \multicolumn{11}{c}{\dss{}} \\
        % \midrule
        Original
          & 83.9 & 28.9 & 71.6 & 43.3 & 56.9
          & 5042 & 15956 & 8202 & 11510 & 10177 \\
        $T_{10240}$ & 82.7 & 26.9 & 73.1 & 44.1 & 56.7 & 2056 & 5458 & 3036 & 3405 & 3489 \\
        $T_{8192}$ & 81.8 & 24.8 & 70.9 & 43.9 & 55.35 & 1795 & 4465 & 2560 & 2841 & 2915 \\
        $T_{7168}$ & 81.8 & 23.3 & 68.6 & 43.0 & 54.18 & 1553 & 3726 & 2251 & 2323 & 2463 \\
        $T_{6144}$ & 80.9 & 20.2 & 66.2 & 42.1 & 52.35 & 1351 & 2821 & 1917 & 1947 & 2009 \\
        $T_{4096}$ & 77.7 & 19.2 & 62.2 & 38.5 & 49.4 & 1054 & 2481 & 1484 & 1564 & 1646 \\
        $T_{2048}$ & 73.2 & 15.8 & 56.9 & 35.9 & 45.45 & 721 & 1029 & 936 & 1084 & 943 \\
        $\text{Group}_{\alpha=0.4}$
          & 74.6 & 25.0 & 69.2 & 43.1 & 53.0
          & 1069 & 4747 & 2162 & 2536 & 2629 \\
       $\text{Group}_{\alpha=0.2}$
          & 78.1 & 28.1 & 68.0 & 44.4 & 54.7
          & 1135 & 5628 & 2635 & 2944 & 3085 \\
       $\text{Group}_{\alpha=0.1}$
          & 77.0 & 29.0 & 69.5 & 44.9 & 55.1
          & 1228 & 6301 & 2808 & 3271 & 3402 \\
      $\text{Group}_{\alpha=0.05}$
          & 74.4 & 30.2 & 65.5 & 43.1 & 53.3
          & 1193 & 4839 & 2457 & 2703 & 2798 \\
      L1-Max-1024
          & 76.4 & 15.0 & 59.4 & 39.1 & 47.5
          & 661 & 1303 & 933 & 938 & 959 \\
        L1-Max-4096
          & 79.7 & 20.0 & 65.0 & 41.0 & 51.4
          & 875 & 1718 & 1159 & 1229 & 1245 \\
        $\text{\mone}_{L_T=2048}$
          & 83.6 & 29.2 & 71.6 & 44.1 & 57.1
          & 1913 & 4815 & 2493 & 2767 & 2895 \\
         $\text{\mone}_{L_T=4096}$
          & 83.9 & 31.0 & 74.1 & 45.7 & 58.7
          & 1914 & 5915 & 3136 & 3579 & 3636 \\
         $\text{\mone}_{L_T=8192}$
          & 85.6 & 31.5 & 75.9 & 47.7 & 60.2
          & 2736 & 6589 & 4162 & 4547 & 4509 \\
         $\text{\mtwo}_{L_T=1024}$
          & 83.0 & 30.6 & 72.8 & 43.7 & 57.5
          & 1362 & 4991 & 2556 & 2837 & 2862 \\
         $\text{\mtwo}_{L_T=2048}$
          & 82.2 & 31.0 & 73.3 & 46.2 & 58.2
          & 1623 & 5158 & 2572 & 2960 & 3059 \\
         $\text{\mtwo}_{L_T=4096}$
          & 84.2 & 34.2 & 75.3 & 47.3 & 60.3
          & 1872 & 5750 & 2981 & 3474 & 3520 \\
         $\text{\mthree}_{L_T=1024}$
          & 82.1 & 33.8 & 72.2 & 43.7 & 58.0
          & 1350 & 4794 & 2254 & 2654 & 2763 \\
         $\text{\mthree}_{L_T=2048}$
          & 83.9 & 31.5 & 75.3 & 46.4 & 59.3
          & 1456 & 5263 & 2679 & 2971 & 3092 \\
         $\text{\mthree}_{L_T=4096}$
          & 83.5 & 35.0 & 73.3 & 46.0 & 59.5
          & 1949 & 5789 & 3080 & 3488 & 3577 \\
      \midrule
      \bottomrule
      \end{NiceTabular}%
    }
    \label{tab:main_results_full}
  \end{table}
  
\section{Full Experimental Results on Out-of-Domain Benchmarks}
\label{sec:ood_full}

\begin{figure}[htbp]
    \centering
    \begin{subfigure}[b]{0.48\textwidth}
        \centering
        \includegraphics[width=\textwidth]{figures/ood/performance_gpqa.pdf}
        \caption{GPQA}
    \end{subfigure}
    \hfill
    \begin{subfigure}[b]{0.48\textwidth}
        \centering
        \includegraphics[width=\textwidth]{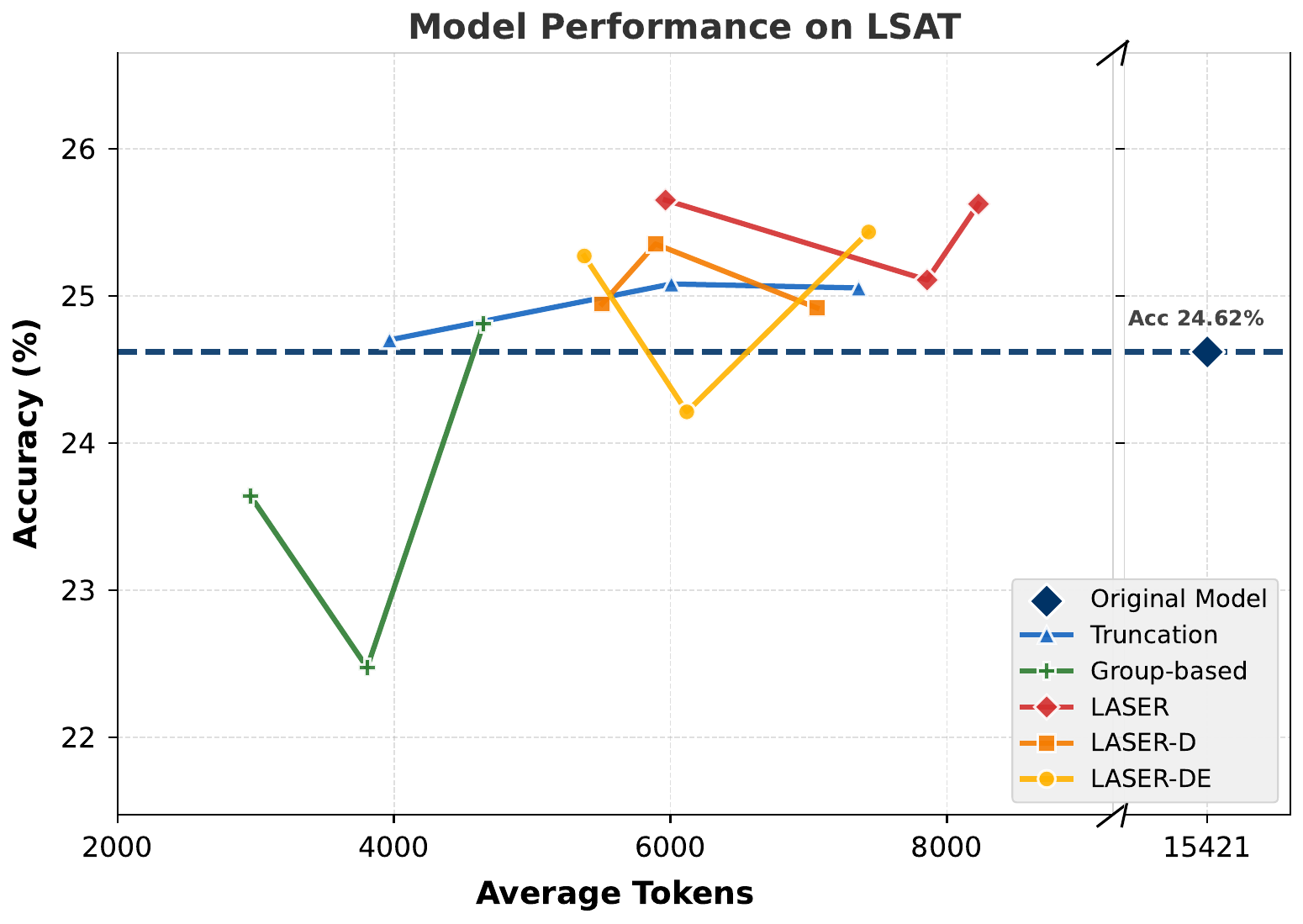}
        \caption{LSAT}
    \end{subfigure}
    
    \vspace{0.5cm}
    
    \begin{subfigure}[b]{0.48\textwidth}
        \centering
        \includegraphics[width=\textwidth]{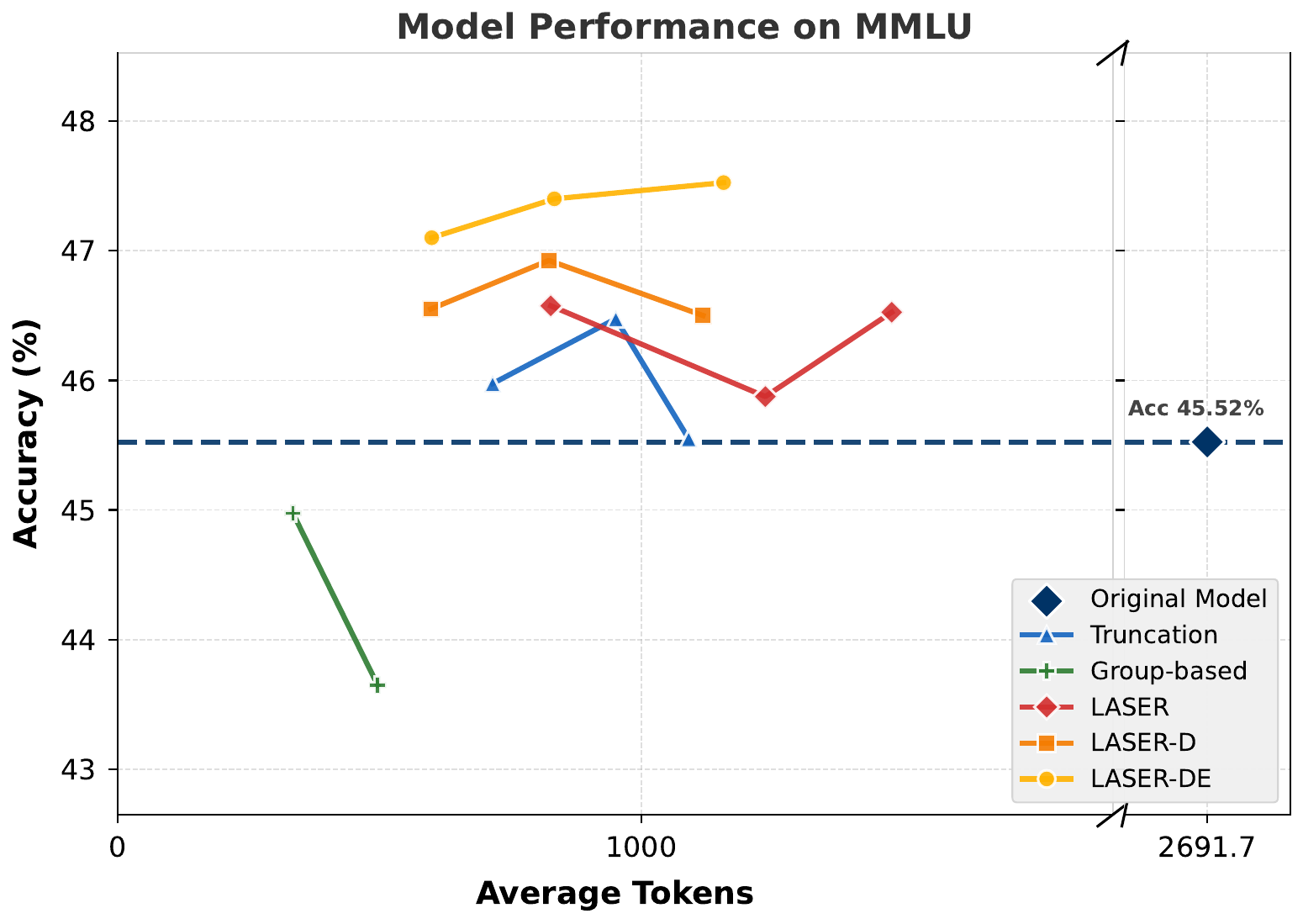}
        \caption{MMLU}
    \end{subfigure}
    \hfill
    \begin{subfigure}[b]{0.48\textwidth}
        \centering
        \includegraphics[width=\textwidth]{figures/ood/performance_average.pdf}
        \caption{Average}
    \end{subfigure}
    
    \caption{Performance on out-of-domain benchmarks including GPQA \citep{rein2023gpqagraduatelevelgoogleproofqa}, LSAT \citep{zhong2021arlsat,wang2022lsat}, and MMLU \citep{hendrycks2021measuringmassivemultitasklanguage}.}
    \label{fig:ood_performance_full}
\end{figure}

Figure~\ref{fig:ood_performance_full} illustrates the performance of various methods on out-of-domain benchmarks, including GPQA \citep{rein2023gpqagraduatelevelgoogleproofqa}, LSAT \citep{zhong2021arlsat,wang2022lsat}, and MMLU \citep{hendrycks2021measuringmassivemultitasklanguage}. Across all benchmarks, \mone, \mtwo{} and \mthree{} consistently demonstrate significant improvements in both accuracy and efficiency. Notably, these improvements extend even to the knowledge-intensive MMLU benchmark, highlighting the robust generalization capabilities of our proposed methods.

\section{Visualization Details}
\label{ap:vis}
In this appendix, we provide details about the visualization of different reward functions depicted in Table~\ref{tab:unified_reward}. These visualizations illustrate how different methods calculate rewards based on response length.

\subsection{Visualization Parameters}

Each visualization captures the relationship between response length and reward value with the following specifications:

\begin{itemize}
    \item \textbf{X-axis}: L(y) represents the response length, ranging from 0 to 20 tokens.
    \item \textbf{Y-axis}: Reward value, with different ranges depending on the method.
    \item \textbf{Line styles}: Solid lines represent rewards for correct responses (\textcolor[rgb]{0.28, 0.39, 0.70}{blue}), while dashed lines represent rewards for incorrect responses (\textcolor[rgb]{0.66, 0.24, 0.19}{red}).
    \item \textbf{Target length} ($L_T$): Set to 10 tokens for all methods.
\end{itemize}

The visualizations were generated using a high-resolution grid of 400 points between 0 and 20 tokens.

\subsection{Unified Reward Formulation}

Each method can be represented using the unified reward formula:
\begin{equation*}
    \hat{R}(x, y) = C(y) + \lambda(y)\cdot S(y)
\end{equation*}

We implement the specific components for each method in this simulation as follows. Note that the paramters are only used for better visualization which are different from the practical experiments.

\paragraph{Vanilla Truncation}
\begin{align*}
    C(y) &= 0 \\
    \lambda(y) &= 1 \\
    S(y) &= \begin{cases}
      R(x,y) & \text{if } L(y) \leq L_T \\
      \rho & \text{if } L(y) > L_T
    \end{cases}
\end{align*}
where $L_T = 10$ and $\rho=0$.

\paragraph{ThinkPrune}
\begin{align*}
    C(y) &= 0 \\
    \lambda(y) &= 1 \\
    S(y) &= \begin{cases}
      R(x,y) & \text{if } L(y) \leq L_A \\
      \rho & \text{if } L(y) > L_A
    \end{cases}
\end{align*}
where $L_A \in \{10, 7.5, 5\}$.

\paragraph{Efficient Reasoning}
\begin{align*}
    C(y) &= R(x,y) \\
    \lambda(y) &= \mathbb{I}(R) \\
    S(y) &= -\alpha \cdot \sigma\left(\frac{L(y) - Mean(y)}{STD(L)}\right)
\end{align*}
where $\mu = 10$ and $\sigma = 2$.

\paragraph{Kimi-k1.5}
\begin{align*}
    C(y) &= R(x,y) \\
    \lambda(y) &= 1 \\
    S(y) &= \begin{cases}
          0.5-\tfrac{L(y)-L_{\min}}{L_{\max}-L_{\min}} & \text{if } \mathbb{I}(R) = 1 \\
          \min\!\left(0,\;0.5-\tfrac{L(y)-L_{\min}}{L_{\max}-L_{\min}}\right) & \text{if } \mathbb{I}(R) = 0
        \end{cases}
\end{align*}
where $L_{min} = 2.5$ and $L_{max} = 20$.

\paragraph{L1-Exact}
\begin{align*}
    C(y) &= R(x,y) \\
    \lambda(y) &= 1 \\
    S(y) &= -\alpha\cdot |L(y) - L_T|
\end{align*}
where $\alpha = 0.03$ and $L_T = 10$.

\paragraph{L1-Max}
\begin{align*}
    C(y) &= 0 \\
    \lambda(y) &= \mathbb{I}(R) \\
    S(y) &= \operatorname{clip}(\alpha \cdot (L(y) - L_T) + \delta, 0, 1)
\end{align*}
where $\alpha = 0.03$ and $L_T = 10$.

\paragraph{\mone}
\begin{align*}
    C(y) &= R(x,y)  \\
    \lambda(y) &= \mathbb{I}(R) \\
    S(y) &= \alpha \cdot \mathbb{I}(L(y)<L_T)
\end{align*}
where $L_T = 10$.

\paragraph{\mtwo}
\begin{align*}
    C(y) &= R(x,y)  \\
    \lambda(y) &= \mathbb{I}(R) \\
    S(y) &= \alpha \cdot \mathbb{I}(L(y)< L_A)
\end{align*}
where $L_A \in \{10, 7.5, 5\}$.

\paragraph{\mthree}
\begin{align*}
    C(y) &= R(x,y)  \\
    \lambda(y) &= 1 \\
    S(y) &= \alpha\cdot\mathbb{I}(R)\cdot\mathbb{I}(L(y) \leq L_A)
    + \alpha\cdot(1-\mathbb{I}(R))\cdot\mathbb{I}(L(y) > L_A)
\end{align*}
where $L_A \in \{12.5, 10, 7.5\}$.

For methods with multiple adaptive target lengths $L_A$ values (ThinkPrune, \mtwo, and \mthree), different shades of the base colors were used:

\begin{itemize}
  \item Correct responses (blue): 
    \textcolor[RGB]{26, 71, 142}{RGB(26,71,142)}, 
    \textcolor[RGB]{62, 101, 184}{RGB(62,101,184)}, 
    \textcolor[RGB]{125, 154, 230}{RGB(125,154,230)}
  \item Incorrect responses (red): 
    \textcolor[RGB]{139, 0, 0}{RGB(139,0,0)}, 
    \textcolor[RGB]{183, 50, 40}{RGB(183,50,40)}, 
    \textcolor[RGB]{224, 93, 86}{RGB(224,93,86)}
\end{itemize}

% \section{Implementations of Budget-Forcing}
% \label{ap:budget_forcing}
% We follow the budget-forcing implementations of \citet{muennighoff2025s1simpletesttimescaling,hou2025thinkprunepruninglongchainofthought}. Specifically, we follow their implementations and modify the codebase of Qwen-Math-Eval. We stop the thinking process of LRMs by appending ``</think>\verb|\n\n|**Final Answer.**''. Since empirically, \dss{} typically summarize its final answer starting with ``\verb|\n\n|**Final Answer.**''. We use the same settings as our evaluations where we sample responses for multiple times with $\operatorname{temperature}=0.6$.

\section{Analysis of Reasoning Behaviros}
\label{ap:behaviors}
\begin{figure}[htbp]
  \centering
  \includegraphics[width=0.8\textwidth]{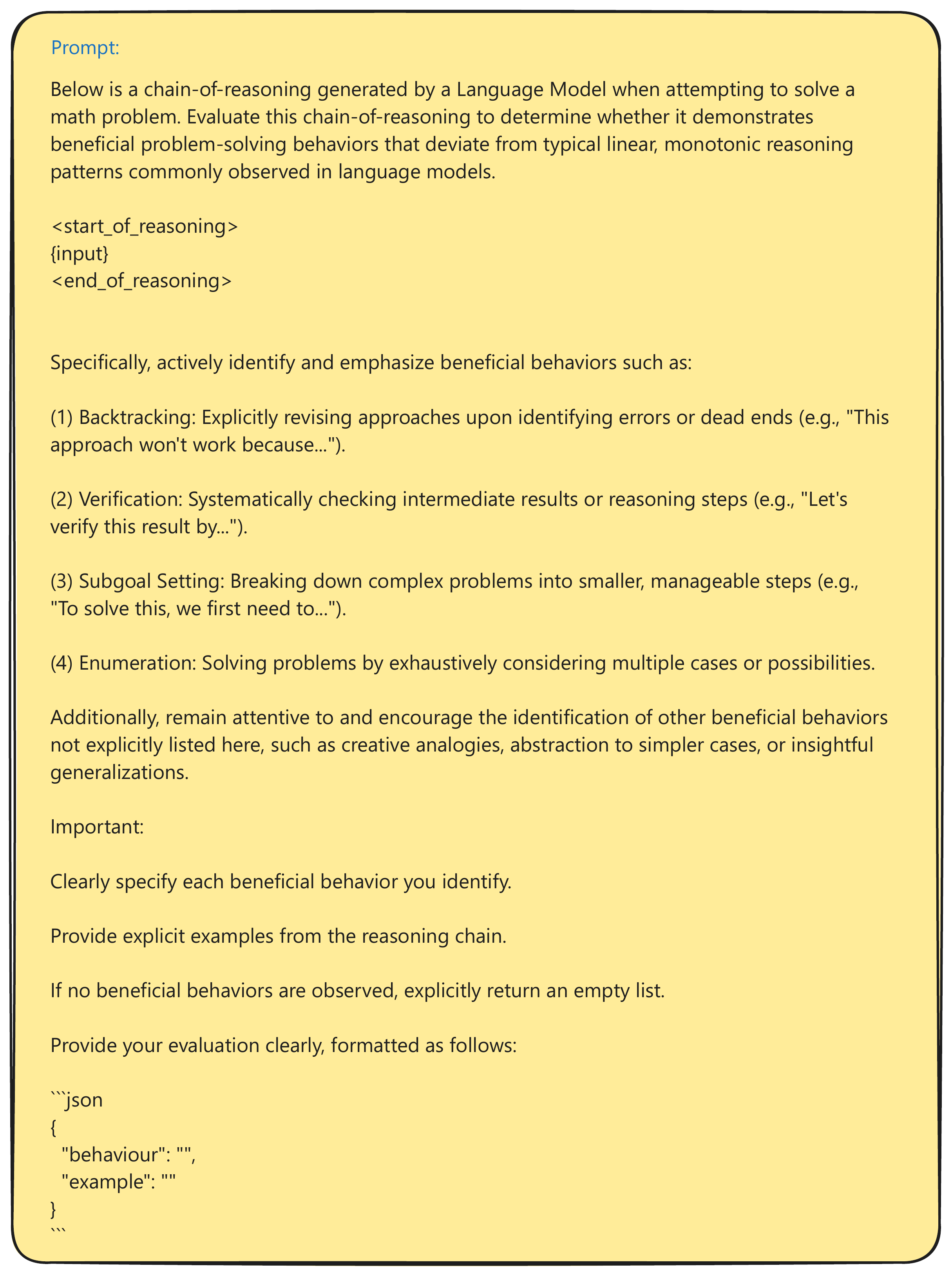}
  \caption{Prompt used to identify and analyze reasoning behaviors with \texttt{gpt-4.1-mini}}
  \label{fig:behavior_prompt}
\end{figure}

We apply the cognitive behavior framework proposed by~\citet{gandhi2025cognitivebehaviorsenableselfimproving} to conduct a detailed analysis of how reasoning behaviors change during our long-to-short RL. We use \texttt{gpt-4.1-mini} to perform a more fine-grained analysis of cognitive behaviors throughout the training process. Following~\citet{zeng2025simplerlzooinvestigatingtamingzero}, we use the prompt shown in Figure~\ref{fig:behavior_prompt} to prompt \texttt{gpt-4.1-mini} to identify and analyze reasoning behaviors. We analyze these behaviors on AIME2024 by sampling one question 16 times, resulting in 480 responses for analysis. Since we start from a LRM, reasoning behaviors such as backtracking naturally appear in every response, especially for challenging benchmarks. We specifically track four key behaviors: \emph{Backtracking}, \emph{Verification}, \emph{Enumeration}, and \emph{Subgoal Setting}. For each behavior, we calculate its frequency ratio relative to all behaviors and report how these ratios change throughout the training process. The complete list of all reasoning behaviors analyzed is provided in Table~\ref{tab:all_behaviors}.

\begin{table}[htbp]
  \centering
  \caption{Complete list of reasoning behaviors}
  \label{tab:all_behaviors}
  \begin{tabular}{c}
  \toprule
  \textbf{Reasoning Behavior} \\
  \midrule
  Subgoal Setting \\
  Enumeration \\
  Verification \\
  Backtracking \\
  Creative Analogy and Abstraction \\
  Abstraction and Parametrization \\
  Analytical Insight via Asymptotic Analysis \\
  Creative Abstraction / Coordinate Setup \\
  Use of Multiple Mathematical Tools and Identities \\
  Creative Analogies and Insightful Generalizations \\
  Algebraic Manipulation and Insightful Generalization \\
  Abstraction to Modular Arithmetic and Divisibility \\
  Creative Analogies and Abstractions \\
  Insightful Generalization / Alternative Modeling \\
  \bottomrule
  \end{tabular}
  \end{table}

\section{Qualitative Analysis on Efficient Reasoning}
\label{ap:case_study}
\begin{figure}[htbp]
  \centering
  \includegraphics[width=0.7\textwidth]{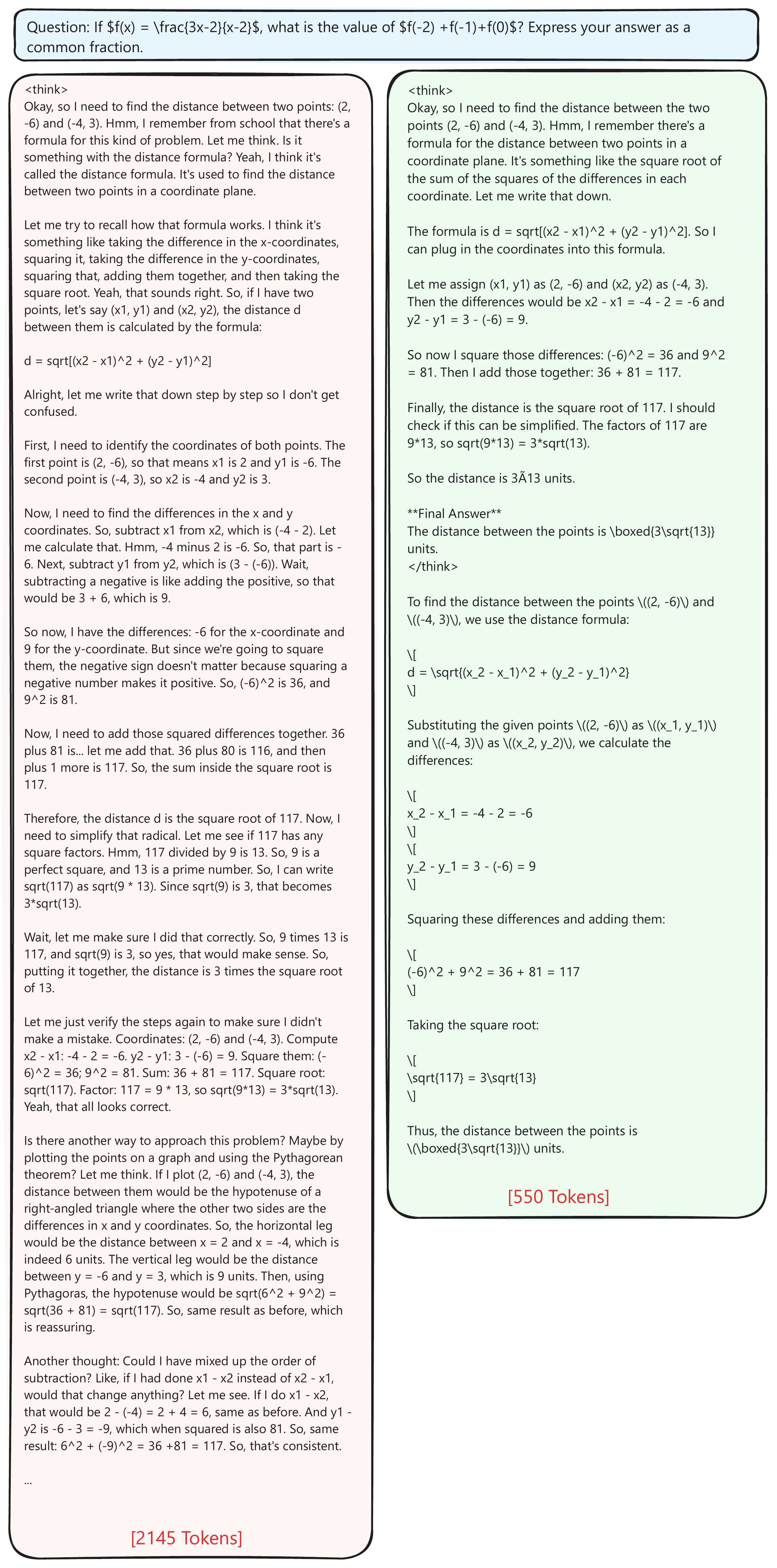}
  \caption{The full example of Figure~\ref{fig:pareto}}
  \label{fig:full_case_study}
\end{figure}

\begin{figure}[htbp]
  \centering
  \includegraphics[width=0.65\textwidth]{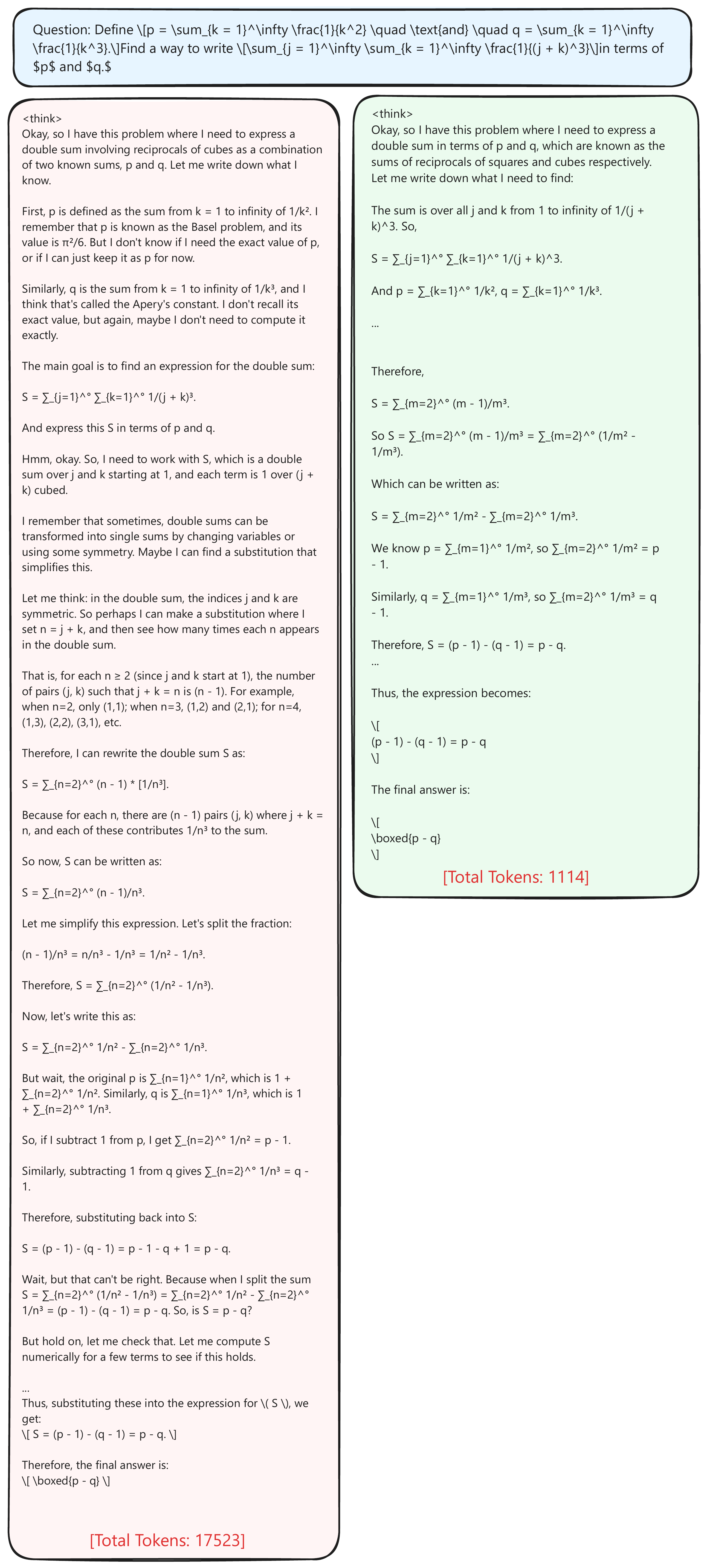}
  \caption{Additional case study demonstrating the evolution of reasoning efficiency. In this example, the original model required over 17K tokens to solve a question from the MATH500 dataset, while our trained model accomplished the same task using only 1K+ tokens.}
  \label{fig:full_case_study_2}
\end{figure}

\begin{figure}[htbp]
  \centering
  \includegraphics[width=0.75\textwidth]{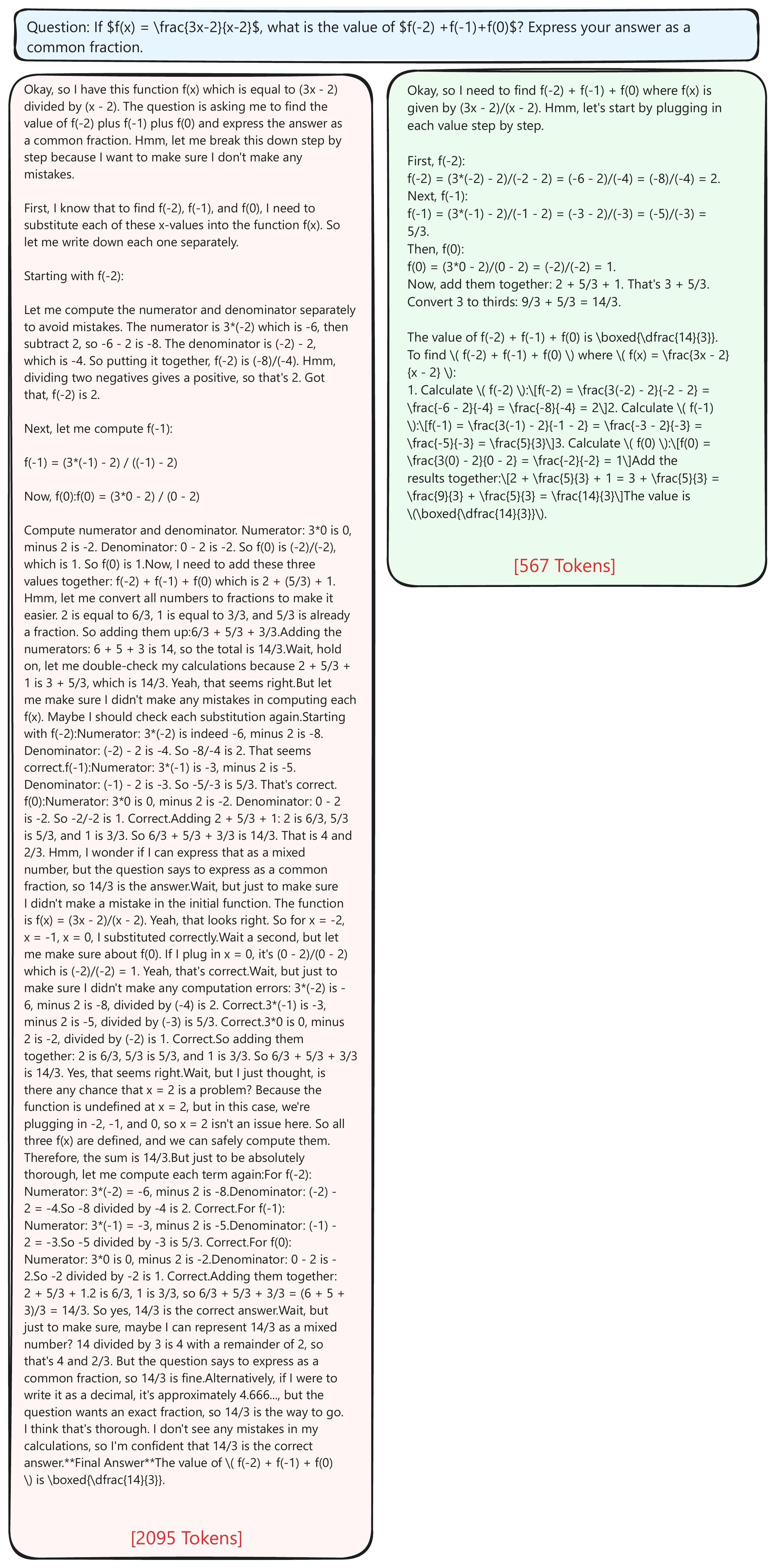}
  \caption{Further example demonstrating improvements in reasoning approach}
  \label{fig:full_case_study_3}
\end{figure}

We conduct a qualitative analysis on the trivial question ``1+1=?'' and the MATH500 dataset to understand how RL improves reasoning efficiency. Comparing the original \dss{} model with the \mtwo-trained version, Figure~\ref{fig:pareto} illustrates how the original model generates repetitive ``self-reflection'' even for trivial questions, while the trained model directly provides the answer. Our analysis of MATH500 (detailed in Figure~\ref{fig:full_case_study}, Figure~\ref{fig:full_case_study_2} and Figure~\ref{fig:full_case_study_3}) reveals that the original model tends towards verbose, redundant explanations of single ideas. In contrast, the \mtwo-trained model expresses the same concepts more succinctly using structured formulas, significantly improving token efficiency. This suggests our RL-based approach not only reduces unproductive backtracking but also encourages a shift towards more concise and direct expression.

\section{Limitations}
\label{ap:limitations}
Despite our work's effective improvements in performance and efficiency, limitations remain. Our and most previous works focus primarily on the math as it provides an excellent verification environment and testbed for validating new methodologies. We believe further validation in code generation and agentic tasks would be valuable to determine if similar favorable trade-offs can be achieved in these contexts. Importantly, our methods were not specifically designed for mathematical tasks but were developed as domain-agnostic approaches that should naturally extend to other areas. In future work, we plan to explore more realistic scenario tasks, particularly those involving agentic reasoning, to further validate our approach and improve the efficacy-efficiency trade-off in more broader areas.

% \newcolumntype{Y}{>{\raggedright\arraybackslash}X}
% \begin{table}[p]
%   \centering
%   \setlength{\tabcolsep}{10pt}
%   \begin{tabularx}{\textwidth}{|Y|Y|}
%       \hline
%       \multicolumn{2}{|c|}{%
%           \cellcolor{gray!20}\large\textbf{Question}
%       } \\
%       \multicolumn{2}{|p{\dimexpr\textwidth-2\tabcolsep-2\arrayrulewidth}|}{%
%       Question: If $f(x) = \frac{3x-2}{x-2}$, what is the value of $f(-2) +f(-1)+f(0)$? Express your answer as a common fraction.
%       } \\
%       \hline
%       \cellcolor{gray!10}\textbf{Response from Original Model} & 
%       \cellcolor{gray!10}\textbf{Response after Training} \\
%       \hline
%       \begin{minipage}[t]{\linewidth}
%           \vspace{0.5ex}
%           Enter the original model's response here...
%           \vspace{0.5ex}
%       \end{minipage}
%       &
%       \begin{minipage}[t]{\linewidth}
%           \vspace{0.5ex}
%           Enter the response after training here...
%           \vspace{0.5ex}
%       \end{minipage}
%       \\
%       \hline
%   \end{tabularx}
%   \caption{Model Response Comparison}
% \end{table}
% \newpage

% \input{checklist}

\end{document}